\documentclass[final]{cvpr}

\usepackage{times}
\usepackage{epsfig}
\usepackage{graphicx}
\usepackage{amsmath}
\usepackage{amssymb}

\usepackage{graphbox}
\usepackage{enumitem}
\usepackage{tabularx}
\usepackage{multirow}
\usepackage{booktabs}
\usepackage{xcolor}
\usepackage[T1]{fontenc}
\usepackage{appendix}

\usepackage{fdsymbol}

\definecolor{limegreen}{rgb}{0.2, 0.8, 0.2}
\definecolor{capri}{rgb}{0.0, 0.75, 1.0}

\newcommand*{\diff}{\color{red}}

\usepackage[pagebackref=true,breaklinks=true,colorlinks,bookmarks=false]{hyperref}

\def\cvprPaperID{****} %
\def\confYear{CVPR 2021}

\begin{document}

\title{BasisNet: Two-stage Model Synthesis for Efficient Inference}

\author{Mingda Zhang\textsuperscript{1}\thanks{This work was done when Mingda Zhang was an intern at Google.} \quad
Chun-Te Chu\textsuperscript{2} \quad
Andrey Zhmoginov\textsuperscript{2} \quad
Andrew Howard\textsuperscript{2} \quad
Brendan Jou\textsuperscript{2} \\
Yukun Zhu\textsuperscript{2} \quad
Li Zhang\textsuperscript{2} \quad
Rebecca Hwa\textsuperscript{1} \quad
Adriana Kovashka\textsuperscript{1}
\\[0.5em]
\textsuperscript{1}Department of Computer Science, University of Pittsburgh \quad
\textsuperscript{2}Google Research\\
\tt{\small{\{mzhang,hwa,kovashka\}@cs.pitt.edu \quad \{ctchu,azhmogin,howarda,bjou,yukun,zhl\}@google.com}}
}

\maketitle

\begin{abstract}
In this work, we present BasisNet which combines recent advancements in efficient neural network architectures, conditional computation, and early termination in a simple new form. Our approach incorporates a lightweight model to preview the input and generate input-dependent combination coefficients, which later controls the synthesis of a more accurate specialist model to make final prediction. The two-stage model synthesis strategy can be applied to any network architectures and both stages are jointly trained. We also show that proper training recipes are critical for increasing generalizability for such high capacity neural networks.
On ImageNet classification benchmark, our BasisNet with MobileNets as backbone demonstrated clear advantage on accuracy-efficiency trade-off over several strong baselines. %
Specifically, BasisNet-MobileNetV3 obtained 80.3\% top-1 accuracy with only 290M Multiply-Add operations, halving the computational cost of previous state-of-the-art without sacrificing accuracy.
With early termination, the average cost can be further reduced to 198M MAdds while maintaining accuracy of 80.0\% on ImageNet.
\end{abstract}

\vspace{-1.5em}
\section{Introduction}
\label{sec:intro}

\begin{table}[t]
\footnotesize
\begin{tabularx}{\linewidth}{>{\hsize=2.0\hsize}X 
  >{\hsize=0.5\hsize\centering\arraybackslash}X
  >{\hsize=0.5\hsize\centering\arraybackslash}X 
}
\toprule
                              & MAdds (FLOPs) & Top-1 Acc. (\%) \\
\midrule
MobileNetV2 1.0x~\cite{sandler2018mobilenetv2} & 300M  & 72.0          \\
CondConv-MobileNetV2 1.0x~\cite{yang2019condconv}$\textcolor{black}{\spadesuit}$     & 329M  & 74.6          \\
DY-MobileNetV2 1.0x~\cite{chen2020dynamic}$\textcolor{black}{\spadesuit}$ & 313M  & 75.2          \\
MobileNetV3-Large~\cite{howard2019searching} & 219M  & 75.2          \\
Dy-MobileNetV3-Large~\cite{zhang2020dynet}          & 228M  & 77.1          \\
ShuffleNetV2 1.5x~\cite{ma2018shufflenet} & 299M  & 72.6          \\
EfficientNet-B0~\cite{tan2019efficientnet} & 390M  & 77.1          \\
EfficientNet-B0 (Noisy Stdt.)~\cite{xie2019self}$\textcolor{capri}{\vardiamondsuit}\textcolor{red}{\varheartsuit}$ & 390M  & 78.1  \\
EfficientNet-B0 (AA + KD)~\cite{wei2020circumventing}$\textcolor{capri}{\vardiamondsuit}\textcolor{black}{\spadesuit}$ & 390M & 78.0 \\
CondConv-EfficientNet-B0~\cite{yang2019condconv}$\textcolor{black}{\spadesuit}$ & 413M  & 78.3          \\
ProxylessNas~\cite{cai2018proxylessnas}                  & 320M  & 74.6          \\
FBNetV2-L1~\cite{Wan_2020_CVPR}                    & 325M  & 77.2          \\
FBNetV3-A~\cite{dai2020fbnetv3}$\textcolor{limegreen}{\clubsuit}$ & 343M  & 78.0          \\
MnasNet-A1~\cite{tan2019mnasnet}                    & 312M  & 75.2          \\
CondConv-MnasNet-A1~\cite{yang2019condconv}$\textcolor{black}{\spadesuit}$ & 325M  & 76.2          \\
\midrule
EfficientNet-B2~\cite{tan2019efficientnet} & 1.0B  & 80.1          \\
EfficientNet-B1 (Noisy Stdt.)~\cite{xie2019self}$\textcolor{capri}{\vardiamondsuit}\textcolor{red}{\varheartsuit}$ & 700M & 80.2 \\
FBNetV3-E~\cite{dai2020fbnetv3}$\textcolor{limegreen}{\clubsuit}$ & 752M & 80.4 \\
OFA~\cite{cai2019once}$\textcolor{capri}{\vardiamondsuit}$ & 595M  & 80.0          \\
\midrule
\textbf{BasisNet-MV3 (Ours)}$\textcolor{capri}{\vardiamondsuit\textcolor{black}{\spadesuit}}\textcolor{red}{\heartsuit}$ & \textbf{290M} & \textbf{80.3} \\
\textbf{+ Early Termination (Ours)}$\textcolor{capri}{\vardiamondsuit}\textcolor{black}{\spadesuit}\textcolor{red}{\heartsuit}$ & \textbf{\textasciitilde 198M}  & \textbf{\textasciitilde 80.0} \\
\bottomrule
\end{tabularx}
\caption{\label{tbl:compact_comparison}Comparison with other efficient networks on ImageNet. Statistics on referenced baselines are cited from original papers.
Different training strategies are applied, \eg, $\textcolor{capri}{\vardiamondsuit}$ knowledge distillation; $\textcolor{red}{\varheartsuit}$ training with extra data; $\textcolor{black}{\spadesuit}$ custom data augmentation; $\textcolor{limegreen}{\clubsuit}$ AutoML-based learned training recipes.}
\vspace{-1em}
\end{table}

\begin{figure*}[t]
    \centering
    \includegraphics[width=0.96\linewidth]{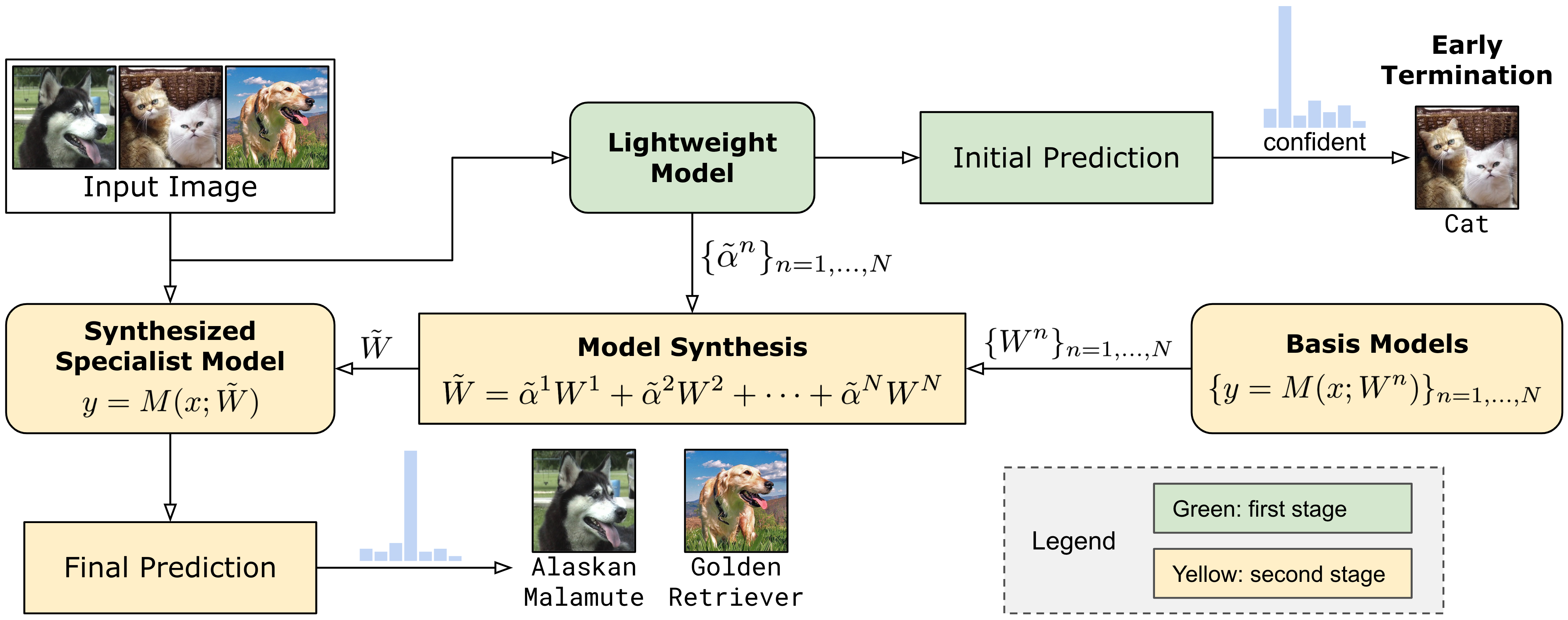}
    \caption{\label{fig:overview}An overview of the BasisNet and more details can be found in Sec.~\ref{sec:basis_models}. For easy images (\eg distinguishing cats from dogs), lightweight model can give sufficiently accurate predictions thus the second stage could be skipped. For more difficult images (\eg, distinguishing different breeds of dogs), a specialist model is synthesized following guidance from lightweight model, which is good at recognizing subtle differences to make more accurate predictions about the given images.}
\end{figure*}

High-accuracy yet low-latency convolutional neural networks enable opportunities for on-device machine learning, and are playing increasingly important roles in various mobile applications, including but not limited to intelligent personal assistants, AR/VR, and real-time speech translations. Therefore designing efficient convolutional neural networks especially for edge devices has received significant research attention. 
Prior research attempted to tackle this challenge from different perspectives, such as novel network architectures~\cite{sandler2018mobilenetv2,zhang2018shufflenet,howard2019searching}, better incorporation with hardware accelerators~\cite{lee2019device}, or conditional computation and adaptive inference algorithms~\cite{bolukbasi2017adaptive,figurnov2017spatially,wang2018skipnet}. 
However, focusing on one perspective \emph{in isolation} may have side effects. 
For example, novel network architectures may introduce custom operators that are not well-supported by hardware accelerators, thus a promising new model may have limited practical improvements on real devices due to a lack of hardware support.
We believe that these perspectives should be better integrated to form a more holistic general approach for broader applicability.

In this paper, we present BasisNet, which takes advantage of progress in all these perspectives and combines several key ideas in a simple new form.
The core idea behind BasisNet is \emph{dynamic model synthesis}, which aims at efficiently generating input-dependent specialist model from a collection of bases, so the resultant model is specialized at handling the given input and can give more accurate predictions.
This concept is flexible and can be applied to any \emph{novel network architectures}.
On the \emph{hardware side}, the two-stage model synthesis strategy allows the execution of the lightweight and synthesized specialist model on different processing units (\eg, CPU, mobile GPUs, dedicated accelerators, \etc) in parallel to better handle streaming data.
The BasisNet design is naturally compatible with \emph{early termination}, and can easily balance between computation budget and accuracy. 
With extensive experiments, we also show that a proper training recipe is critical to mitigate overfitting and improve generalizability.

An overview of the BasisNet is shown in Fig.~\ref{fig:overview}. 
Using image classification as an example, our BasisNet has two stages: the first stage relies on a \emph{lightweight model} to preview the input image and produce both an initial prediction and a group of combination coefficients. In the second stage, the coefficients are used to combine a set of models, which we call \emph{basis models}, into a single one to process the image and generate the final prediction. The second stage could be skipped if the initial prediction is sufficiently confident. The basis models share the same architecture but differ in some weight parameters, while other weights are shared to avoid overfitting and reduce the total model size.

We validated BasisNet with different generations and sizes of MobileNets and observed significant improvements in inference efficiency. 
In Table~\ref{tbl:compact_comparison} we show comparisons
with selected efficient networks on ImageNet classification benchmark. Notably, \emph{without} using early termination, our BasisNet with 16 basis models of MobileNetV3-large only requires 290M Multiply-Adds (MAdds) to achieve 80.3\% top-1 accuracy, halving the computation cost of previous state-of-the-art~\cite{cai2019once} without sacrificing accuracy. If we enable early termination, the \textit{average} cost can be further reduced to 198M MAdds with the top-1 accuracy remaining 80.0\% on ImageNet.\footnote{Average cost is reduced since easy inputs are only handled by lightweight model; max remains 290M MAdds.} 

In summary, our main contribution is two-fold:
\begin{itemize}[nolistsep,noitemsep]
    \item We propose a two-stage model synthesis strategy that combines efficient neural networks, conditional computation, and early termination in a simple new form. Our BasisNet achieves state-of-the-art performance of accuracy with respect to computation budget on ImageNet even \emph{without} early termination; if enabling early termination, the average computation cost can be further reduced with only marginal accuracy drop.
    \item We propose an accompanying training recipe for the new BasisNet, which is critical to improve generalizability for high capacity dynamic neural networks, and can also improve the performance of other models.
\end{itemize}

\section{Related Work}

\paragraph{Efficient neural networks.}
Different approaches for building efficient networks have been studied. Early effort includes knowledge distillation~\cite{hinton2015distilling}, post-training pruning~\cite{han2015deep} and quantization~\cite{courbariaux2016binarized}. Later work distinguishes \emph{model complexity} (size) and \emph{run-time latency} (speed), and optimizes for them either with human expertise~\cite{sandler2018mobilenetv2,zhang2018shufflenet} and/or neural architecture search~\cite{howard2019searching,tan2019efficientnet}. All these approaches aim at producing a \emph{static} model that is generally efficient but agnostic to inputs. On the contrary, our BasisNet is built on efficient network architectures, and is \emph{dynamically} adaptive based on inputs. In this work we optimize for inference speed rather than model size.

\vspace{-1em}
\paragraph{Conditional computation.}
Several prior work have explored accelerating inference by skipping part of computation graph based on input-dependent signals.
For example, \cite{figurnov2017spatially} propose a ResNet extension that dynamically adjusts the number of executed layers based on image regions.
\cite{teja2018hydranets} propose HydraNet which creates multiple parallel branches across the network, and adopts a soft gating module to selectively activate few branches to reduce inference cost. 
\cite{shazeer2017outrageously} use mixture of experts with a gating network to choose from thousands of candidates. 
Recently, \cite{yang2019condconv} propose conditionally parameterized convolution (CondConv), which applies weighted combinations of convolution kernels. This idea is adopted by several later work~\cite{zhang2019domain,chen2020dynamic}, because it has equivalent expressive power as linear mixture of experts, but requires much fewer computations than combining feature maps.
However, one common characteristic of these approaches is that their conditioning modules are inserted before each configurable component (e.g., layer or branch), thus these dynamic adjustments only rely on local information. This concept is defined by \cite{chen2019you}, and according to them lacking global knowledge may be less than optimal because shallower layers cannot benefit from semantic knowledge which is only available from deeper layers. 
Some other work also identified similar issues and have attempted to leverage global knowledge in dynamic modulation.
For example, 
in SkipNet~\cite{wang2018skipnet} a gating network is built to conditionally skip certain layers in the backbone, and the authors report that the best performance comes from a RNN-based gating network because it can access feature maps across multiple layers.
\cite{chen2019you} introduce GaterNet where a dedicated deep neural network is used to analyze the inputs before generating input-dependent masks for the filters in backbone network. 
BasisNet use a lightweight but fully-fledged model to process the inputs and produce combination coefficients, thus the model synthesis is relying on semantic-aware global knowledge.
Different from SkipNet and GaterNet, our lightweight model can synthesize new kernels that do not exist beforehand via linear combination. Another distinction is that by separating conditioning model from backbone, our BasisNet is more flexible and easier to adapt to different architectures and hardware constraints.

\vspace{-1em}
\paragraph{Cascading networks and early exiting.}
Since input samples are naturally of varying difficulty, using a single model to equally process all inputs with a fixed computation budget is wasteful. This observation has been leveraged by prior work, e.g., 
the famous Viola-Jones face detector~\cite{viola2001rapid} built a cascade of increasingly more complex classifiers to achieve real-time execution. Similar ideas were also used in deep learning, e.g., reducing unnecessary inference computations for easy cases in a cascaded system~\cite{venkataramani2015scalable}, attaching multiple classification heads on different layers~\cite{teerapittayanon2016branchynet,huang2018multiscale}, or cascading multiple models~\cite{bolukbasi2017adaptive}. One common limitation in previous work is that only the exit point adapts to the samples but the underlying models remain static.
Instead, our BasisNet dynamically adjusts the convolution kernel weights based on the guidance from lightweight model, thus the synthesized specialist can better handle the more difficult cases.

\section{Approach}

In general, our BasisNet has two stages: the first stage lightweight model, and the second stage model synthesis from a set of basis models.
Given a specific input, the lightweight model generates two outputs, an initial prediction and a group of basis \textit{combination coefficients}. If the initial prediction is of high confidence, the input is presumably easy and BasisNet could directly return the initial prediction and terminate early.
But if the initial prediction is less confident (implying the input is difficult, \eg identifying dogs by breed),
the \textit{coefficients} will be used to guide the synthesis of a specialist model in the second stage. The synthesized specialist will handle final prediction. 

\subsection{Lightweight model}
\label{sec:lightweight_model}
The lightweight model is a fully-fledged network 
handling two tasks: generating initial category prediction and generating combination coefficients for second stage model synthesis. The first is a standard classification task thus we only elaborate on the second below. %
Assuming there are $N$ basis models and each has $K$ layers, the lightweight model will predict combination coefficients $\alpha \in \mathcal{R}^{K \times N}$
\begin{gather}\label{eq:1}
    \alpha = \phi(\text{LM}(f(x)))
\end{gather}
where $\text{LM}$ stands for \textit{lightweight model} and $\phi$ represents a non-linear activation function. We use softmax by default because it enforces convexity, which promotes sparsity and can lead to more efficient executions.
$f(x)$ represents a transformation of the input image, and we typically use $f(x) = x$ or $f(x) = \text{DownSampling}(x)$.

\subsection{Basis model synthesis}
\label{sec:basis_models}
Our basis models are a collection of model candidates, which share the same architecture but differ in model parameters. By combining basis models with different weights, a specialist network can be synthesized.
Various strategies can be used for building basis models, such as mixture of experts~\cite{shazeer2017outrageously} or using multiple parameter-efficient patches~\cite{mudrakarta2018k}. We explored a few options and found that the recently proposed CondConv~\cite{yang2019condconv} best fits our needs for building a low-latency but high-capacity model.

Specifically, consider a \emph{regular} deep network with image input $x$. Assume the output of the $k$-th convolutional layer is $O_k(x)$, which could be obtained by
\begin{gather}
    O_k(x) =
    \begin{cases}
        \phi(W_0 * x), & \text{if} ~ k=0\\
        \phi(W_k * O_{k-1}(x)), & \text{if} ~ k>0
    \end{cases}
\end{gather}
where $W_k$ represents the convolution kernel at the $k$-th layer and $*$ represents a convolution operation.
For simplicity some operations like batch normalization and squeeze-and-excitation are omitted from the notation.
In BasisNet, different inputs will be processed by different, input-dependent kernel $\tilde{W}_k$ at $k$-th layer, which is obtained by linearly combining the kernels from $N$ basis models at $k$-th layer, denoted by $\{W_k^n\}_{n=1,\dots,N}$:
\begin{gather}
    \tilde{W}_k = \tilde{\alpha}_k^1 \cdot W_k^1 + \cdots + \tilde{\alpha}_k^N \cdot W_k^N
\end{gather}
where $\tilde{\alpha}_k^n$ represents the weight for the $k$-th layer of the $n$-th basis. We use $\tilde{W}$ and $\tilde{\alpha}$ to emphasize their dependency on $x$. This design allows us to increase model capacity effectively but retain the same number of convolution operations.
Besides, since the number of parameters is much less than number of MAdds in a single basis architecture, the combination only marginally increase the computation cost. 

Besides, using sparse convex coefficients further reduces the overhead.
Thus we generally consider convex coefficients, but also studied two special cases:
\begin{itemize}[nolistsep,noitemsep]    %
    \item $\alpha_k$ is the same for all layers. In this case, the combination is \emph{per-model} instead of \emph{per-layer}.
    \item $\alpha_k$ as an $N$-dimension vector is one-hot encoded. In this case, model synthesis becomes \emph{model selection}.
\end{itemize}

\vspace{-1em}
\paragraph{Key difference from CondConv.}
Our model synthesis mechanism is inspired by CondConv~\cite{yang2019condconv} but there exists many distinctions.
In CondConv the combination coefficients for $k$-th layer are computed following
\begin{gather}
    \alpha_k = \phi(\text{FC}(\text{GAP}(O_{k-1}(x))))
\end{gather}
where FC stands for \textit{fully connected layer} and GAP stands for \textit{global average pooling}.
This formulation shows the dynamic kernels in CondConv can only be synthesized layer by layer, because the combination coefficients for next layer depend on output of previous layer. This complicates scheduling of computation thus is not hardware friendly~\cite{zhang2019domain}.
In BasisNet, the issue is addressed by the lightweight model, which generates the combination coefficients for all layers simultaneously as shown in Equation~\ref{eq:1}. Therefore the entire specialist model can be synthesized all at once. 
Separating kernel combination from execution also enables BasisNet to be easily deployed \emph{to} (or even \emph{across}) different hardware accelerators on edge devices if needed. 
Besides, early termination is naturally supported by BasisNet, but is much harder to be incorporated for CondConv. Arguably one can try attaching additional prediction heads like~\cite{teerapittayanon2016branchynet} to enable ``layer-level early termination'' for CondConv. However, this change requires non-trivial efforts for designing the proper exit points in CondConv, let alone introducing extra computational cost. More specifically, since the backbone efficient network is already highly compact (\eg MobileNets), it is unlikely that early layers can offer signals sufficient for prediction which is required for early termination. For BasisNet, the signal comes from fully-fledged lightweight model, which generates the prediction as a side product thus offers early termination for free.
Lastly, BasisNet is complementary to CondConv, as we find (in Sec.~\ref{sec:condconv}) that combining CondConv and BasisNet can further boost prediction accuracy.

\subsection{Training BasisNet properly}
\label{sec:proper_training}

BasisNet significantly increases model capacity, but the risk of overfitting also increases. We found the standard training procedures used to train MobileNets lead to severe overfitting on BasisNet. Here we describe a few regularization techniques that are crucial for training BasisNet successfully. This is also a key contribution of our work, as previously there is no good practice on how to effectively train such high capacity dynamic neural networks.

\begin{itemize}[nolistsep,noitemsep,leftmargin=*]
    \item \textbf{Basis model dropout (BMD)}
    Inspired by \cite{gastaldi2017shake}, we experimented with randomly shutting down certain basis model candidates during training. 
    It is similar to applying DropConnect~\cite{wan2013regularization} on the predicted coefficient matrix from the lightweight model. 
    We find this approach is extremely effective against ``experts degeneration''~\cite{shazeer2017outrageously} where the controlling model always picks the same few candidates and never activates the rest.
    \item \textbf{AutoAugment (AA)}
    AutoAugment~\cite{cubuk2019autoaugment} is a search-based procedure for finding specific data augmentation policy towards a target dataset. We find that replacing the original data augmentation in MobileNets~\cite{sandler2018mobilenetv2} with the ImageNet policy in AutoAugment can significantly improve the model generalizability.
    \item \textbf{Knowledge distillation}
    \cite{hinton2015distilling} showed that using soft targets from a well-trained teacher network can effectively prevent a student model from overfitting. We observe that knowledge distillation is also effective on training BasisNet, and find EfficientNet-B2 with noisy student training~\cite{xie2019self} can be a good teacher.
\end{itemize}
In addition to stronger regularization, we applied a few other tricks in order to properly train BasisNet. Since the lightweight model directly controls how the specialist model is synthesized, any slight changes in the combination coefficients will propagate to the parameter of the synthesized model and finally affect the final prediction. Since we train the two stages from scratch, this is especially troublesome at the early phase when the lightweight model is still ill-trained. To deal with the unstable training, we introduced $\epsilon \in [0, 1]$ to balance between a uniform combination and a predicted combination coefficients from the lightweight model,
\begin{gather}
    \alpha' = \epsilon \cdot \frac{1}{N} \cdot \mathbf{1}^{K \times N} + (1 - \epsilon) \cdot \alpha
\end{gather}
When $\epsilon=1$ all bases are combined equally while when $\epsilon=0$ the synthesis is following the combination coefficients.
In practice $\epsilon$ linearly decays from 1 to 0 in the early phase of training then remains at 0, thus the lightweight model can gradually take over the control of model synthesis. This approach effectively stabilizes training and accelerates convergence. A recent work~\cite{chen2020dynamic} proposed temperature-controlled softmax to achieve similar goal.

All models in both stages are trained together in an end-to-end manner via back-propagation. In other words, all basis models are trained from scratch by gradients from the synthesized model.
The total loss includes two cross-entropy losses for the synthesized model and the lightweight model, respectively, and L2 regularization,
\begin{equation}
    \begin{split}
    L = &-\log P(y|x; \tilde{W}) + \lambda (-\log P(y|f(x); W_\text{LM})) \\
    &+ \Omega(\{W^n\}_{n=1,\dots,N}, W_{\text{LM}})
    \end{split}
\end{equation}
where $\lambda$ is the weight for cross-entropy loss from lightweight model ($\lambda=1$ in our experiments), and $\Omega(\cdot)$ is a L2 regularizer applied to all model parameters. The lightweight model receives all gradients, while basis models are only updated by the first term and regularization. 

\section{Experiments}

\subsection{Dataset and model architecture setup}
We demonstrate the effectiveness of BasisNet on both MobileNetV2 and MobileNetV3 architectures, and evaluate on the ImageNet ILSVRC 2012 classification dataset~\cite{russakovsky2015imagenet} consisting of 1.28M images for training and 50K for validation. We did not explicitly use extra data, but one teacher model we used for knowledge distillation, \ie, EfficientNet-b2 with noisy student training~\cite{xie2019self}, is obtained with extra data.
For BasisNet-MV2, the basis models follow the architecture described in Table~2 of \cite{sandler2018mobilenetv2}. For simplicity in notation, we sequentially number all the layers starting from L0,
\eg the first \texttt{conv2d} layer is L0 and the \texttt{avgpool 7x7} layer is L19. 
For BasisNet-MV3, the basis models follow the MobileNetV3-large architecture described in Table~1 of \cite{howard2019searching}. We also sequentially number all the layers, \eg the \texttt{pool,7x7} layer is L17.

For fair comparison, we \textit{retrained all models including BasisNet and all the baselines using the same training recipe}, and reported the performance \textit{without} early termination except for Sec.~\ref{sec:early_exiting}. Note that the lightweight model introduces computation overhead for BasisNet, but our reported MAdds statistics for BasisNet always \textit{include} the lightweight model. More details about our model as well as training recipes can be found in supplementary materials.

\begin{figure*}[htbp]
    \centering
    \begin{minipage}{.32\textwidth}
    \centering
    \includegraphics[width=\textwidth]{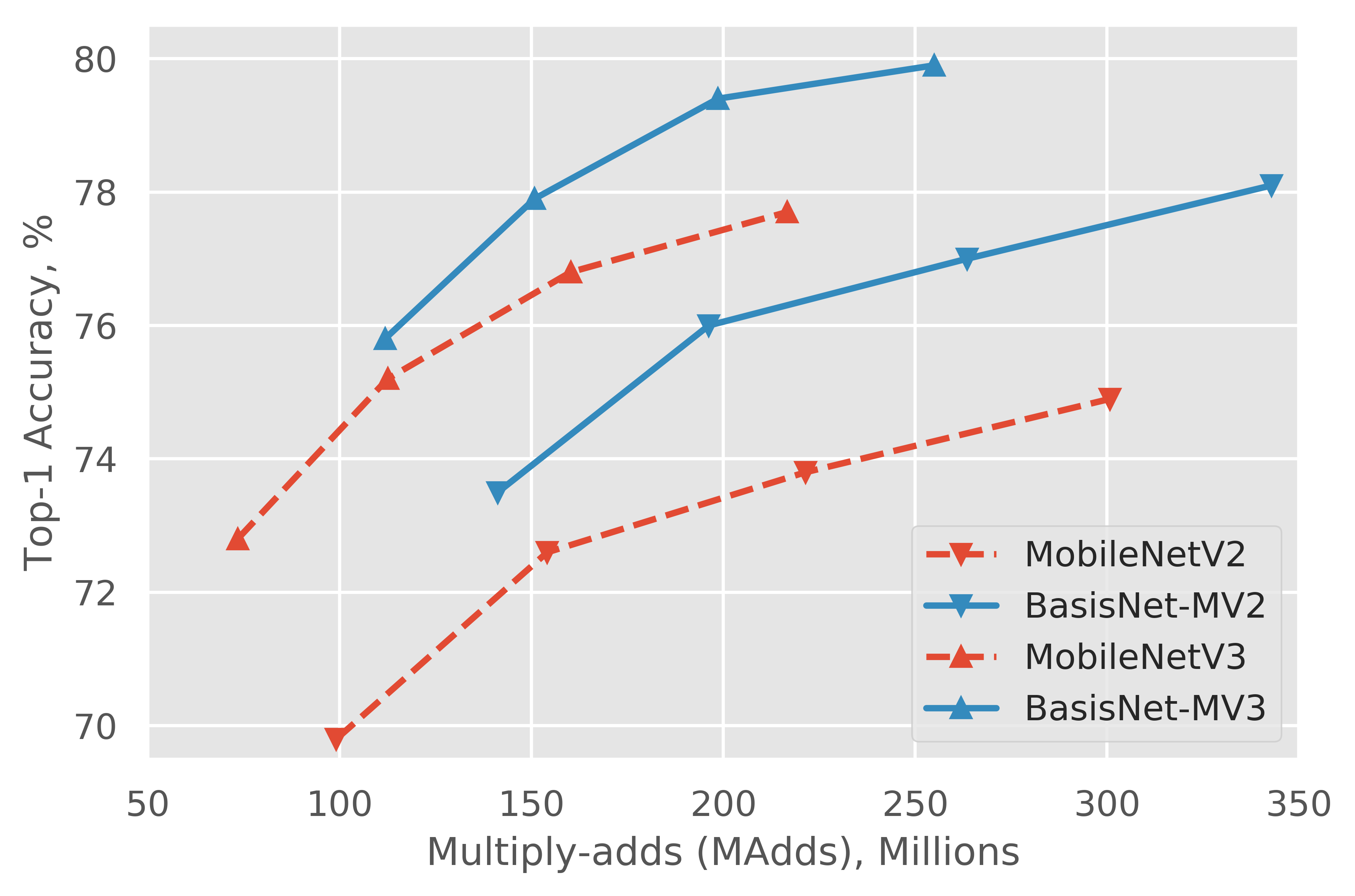}
    \caption{\label{fig:performance}Accuracy-MAdds trade-off comparison of the proposed BasisNet and MobileNet on ImageNet validation set.}
    \end{minipage}%
    \enskip
    \begin{minipage}{0.32\textwidth}
    \centering
      \includegraphics[width=\textwidth]{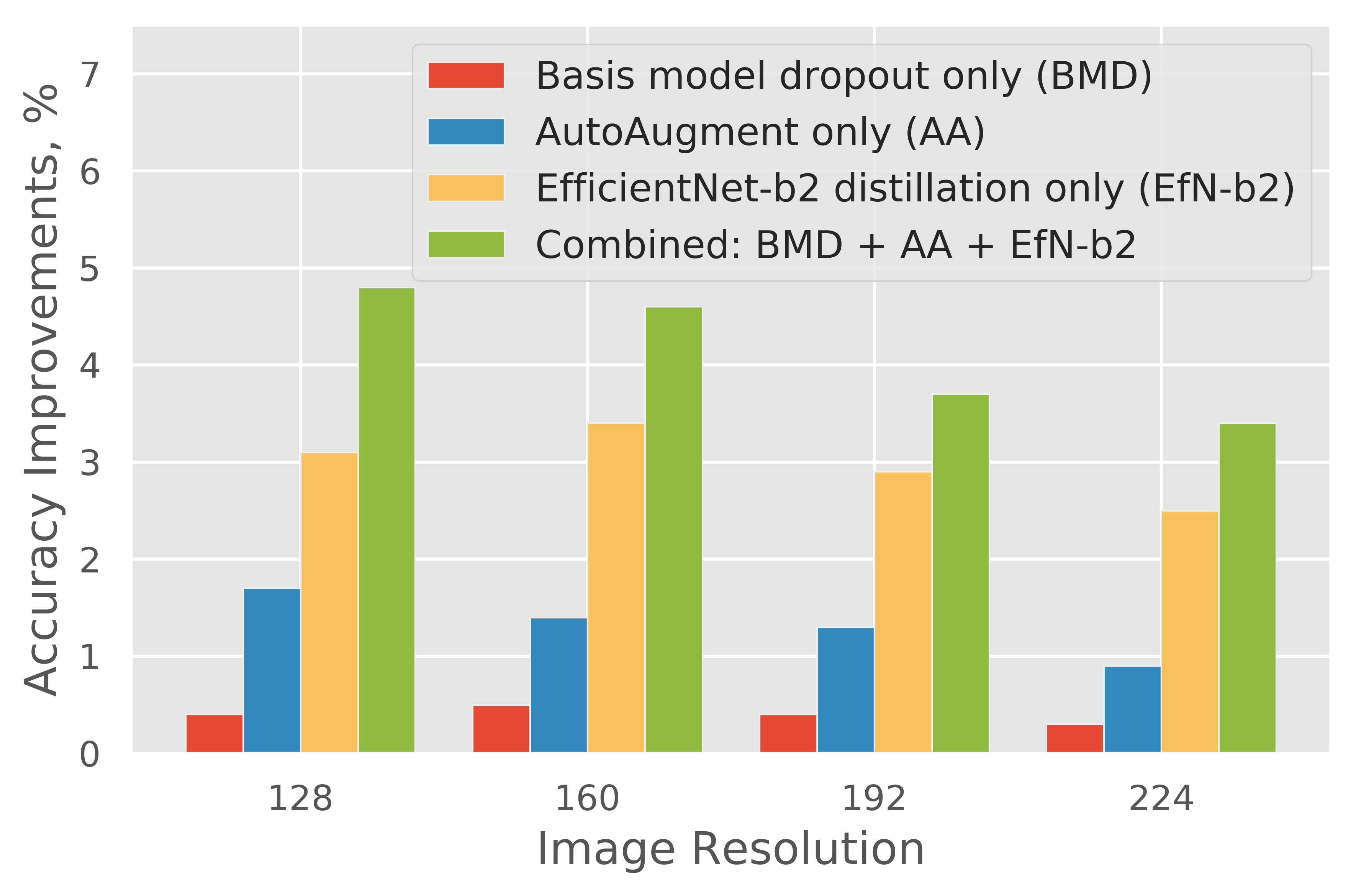}
      \caption{\label{fig:regularization}Performance boost with various regularizations on BasisNet-MV2. All combined gives the largest improvement.}
    \end{minipage}%
    \enskip 
    \begin{minipage}{.32\textwidth}
    \centering
      \includegraphics[width=\textwidth]{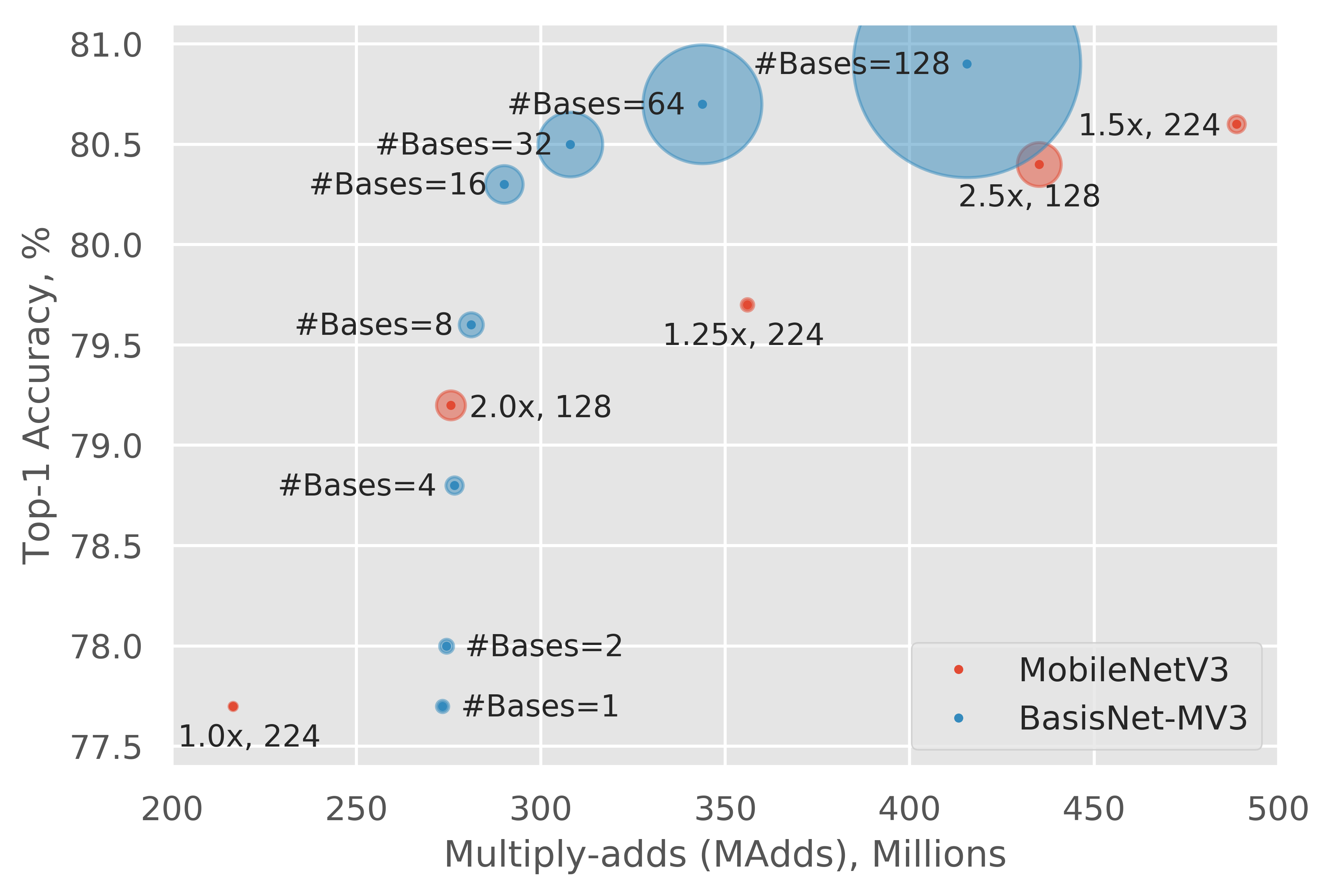}     
      \caption{\label{fig:num_routes}Prediction accuracy monotonically increases when more bases are added to the basis models.}
    \end{minipage}%

\end{figure*}

\begin{figure*}[ht]
    \centering
    \includegraphics[width=0.96\textwidth]{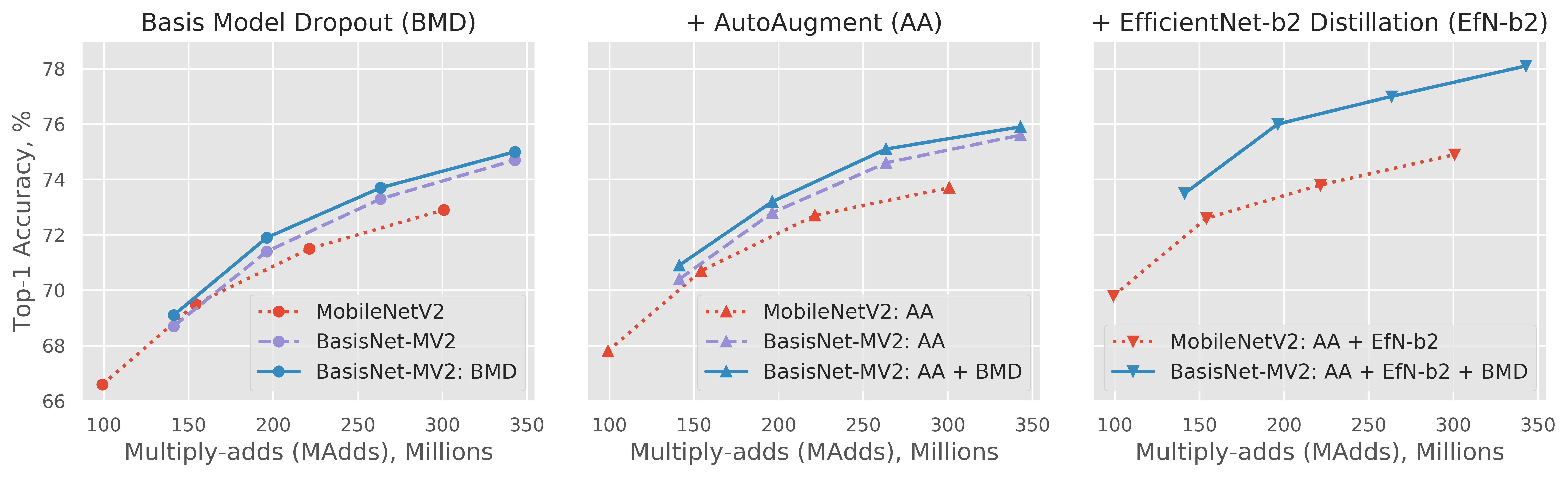}
    \caption{\label{fig:improvements}MobileNet and BasisNet training using different regularizations. BasisNet uses MV2-0.5x as its lightweight model and 8 MV2-1.0x for basis models. Input image resolutions vary from \texttt{\footnotesize\{128, 160, 192, 224\}}. Note that basis model dropout (BMD) is not applicable to MobileNet because it has only one model.}
\end{figure*}

\subsection{Comparison with MobileNets}
\label{sec:comparison_mobilenets}

For both BasisNet-MV2 and BasisNet-MV3, we compute the accuracy-MAdds curves by 
varying the input image resolution to the synthesized model from \texttt{\footnotesize\{128, 160, 192, 224\}}. We compute the curves for the MobileNets in the same way. As shown in Fig.~\ref{fig:performance}, even with the computation overhead of the lightweight model, our BasisNets consistently outperform the MobileNets with large margins.

\subsection{The effect of regularization for proper training}
\label{sec:experiments_proper_training}
In Fig.~\ref{fig:regularization} we show the performance improvements when different regularizations (basis model dropout, AutoAugment, and knowledge distillation) discussed in Sec.~\ref{sec:proper_training} are individually applied to BasisNet-MV2 training, as well as combined altogether. Each regularization helps generalization, and the most effective single regularization is the knowledge distillation. By combining all strategies the validation accuracy increases the most. In fact, we observed that the proposed training recipe also helps improving performance of other models like original MobileNets, as shown in Fig.~\ref{fig:improvements}. 
However, applying the regularization is more crucial for BasisNet training, as the top-1 accuracy of BasisNet-MV2 (1.0x224) improves by $+3.4$ percentage points ($74.7\% \rightarrow 78.1\%$), while for MobileNetV2 the improvement is $+2.0$ percentage points ($72.9\%  \rightarrow 74.9\%$).

\subsection{Number of bases in basis models}
\label{sec:number_of_basis}

We varied the number of bases to investigate their effect on the model size, inference cost and final accuracy. Intuitively, the more bases in the candidate pool, the more diverse domains the final synthesized model can adapt to.
We chose a fix-sized MV3-small (1.0x224) as our lightweight model, and use different numbers of MV3-large (1.0x224) for basis. As shown in Fig.~\ref{fig:num_routes}, the top-1 accuracy improves monotonically with increased number of bases.
With 16 bases, our BasisNet-MV3 achieved 80.3\% accuracy with 290M MAdds. 
The shaded area represents the relative model size (\#Params). Note that we explicitly trained a regular MobileNetV3-large with large multiplier and low image resolution (2.5x128), so it has similar model size with BasisNet. We show that BasisNet requires only 2/3 of computations (290M vs 435M) to achieve the comparable accuracy with the MobileNetV3 counterpart (80.3\% vs 80.4\%).

\subsection{Comparison with CondConv}
\label{sec:condconv}

We re-implemented CondConv\footnote{Our re-implementation of CondConv-MV2 achieved $76.2\%$ accuracy, better than the reported 74.6\% from \cite{yang2019condconv}.} to directly compare with our BasisNet. We choose MobileNetV3 as backbone,
and selected $N=16$ for both BasisNet and CondConv from layers 11 to 15.
We chose MV3-small as the lightweight model for BasisNet, and disabled early termination for fair comparison. All models including CondConv baselines are re-trained using the same recipe as in Sec.~\ref{sec:proper_training}.

\begin{table}[t]
\small
\centering
\begin{tabularx}{\linewidth}{
  >{\hsize=1.3\hsize}X
  >{\hsize=.5\hsize\centering\arraybackslash}X
  >{\hsize=.55\hsize\centering\arraybackslash}X 
  >{\hsize=.65\hsize\centering\arraybackslash}X }
 \toprule
 Model & Activation & MAdds & Top-1 Acc. \\
 \midrule
 CondConv-MV3 & Sigmoid & 253M & 79.9\% \\
 BasisNet-MV3 & Softmax & 290M & 80.3\% \\
 BasisNet-MV3 & Sigmoid & 290M & 80.0\% \\
 (BasisNet+CC)-MV3 & Softmax & 290M & 80.5\% \\
 \bottomrule
\end{tabularx}
\caption{\label{tbl:condconv}Comparison of BasisNet with CondConv.}
\vspace{-1em}
\end{table}

The top-1 accuracy for CondConv-MV3 and BasisNet-MV3 is $79.9\%$ and $80.3\%$ respectively, although BasisNet has relatively larger overhead due to the lightweight model.
However, we find that BasisNet is more flexible than CondConv. CondConv reports that simultaneously activating multiple routes is essential for any single input, therefore sigmoid activation has to be used. For BasisNet, we find both sigmoid and softmax work fine ($80.0\%$ and $80.3\%$ accuracy respectively). In fact, using softmax can lead to sparse and even one-hot combination coefficients (see Sec.~\ref{sec:disturbance}), which may help reducing latency from model loading I/O perspective. %
We also experiment to combine CondConv with BasisNet, and the accuracy can be further boosted to $80.5\%$, showing the performance gain from BasisNet is complementary to CondConv.

\subsection{Early stop to reduce average inference cost}
\label{sec:early_exiting}
\begin{figure}[t]

  \centering
  \includegraphics[width=0.75\linewidth]{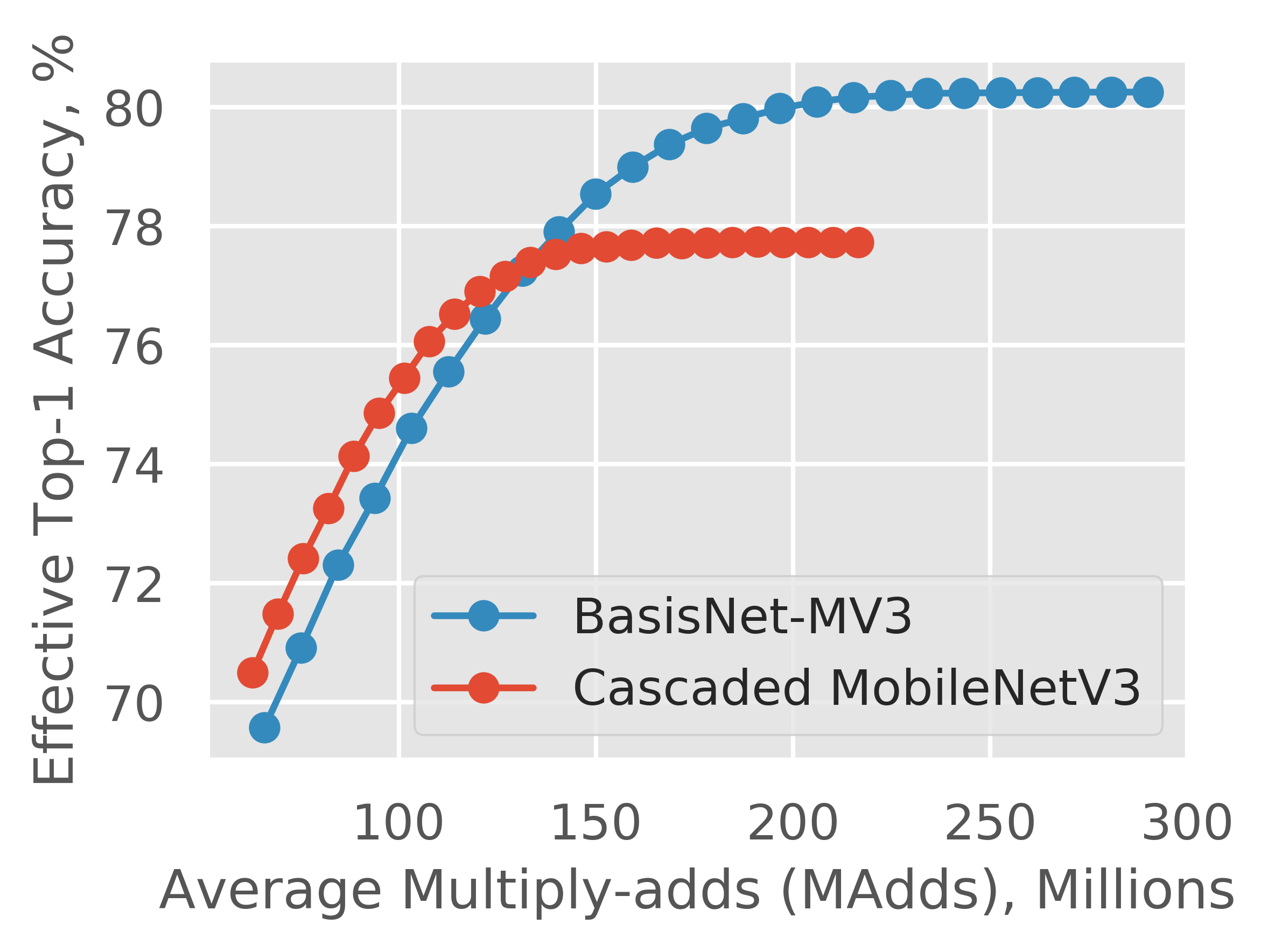}
\caption{\label{fig:early_exiting} Simulated accuracy comparison of BasisNet-MV3 and cascaded MobileNets with early exiting under varying thresholds.}
\end{figure}

The two stage design of BasisNet naturally supports early termination, as the lightweight model can make an initial prediction. We chose the maximum value of softmax probability~\cite{huang2018multiscale} of initial prediction as the criterion to measure the confidence.
Specifically, for each input image if the initial prediction confidence is higher than a predefined threshold then second stage could be skipped, otherwise the second stage specialist needs be synthesized to make the final prediction.

We verified the early termination strategy on ImageNet validation set with a well-trained BasisNet-MV3 (1.0x224,16 basis) model.
In Fig.~\ref{fig:early_exiting}, we alter thresholds of initial prediction confidence and plot the average cost and accuracy of BasisNet. For fair comparison, we cascade two well-trained MobileNets of the same size as the lightweight model and basis model respectively.
In general the figure shows that BasisNet achieves better results for the same cost, except when the computation budget is very limited. Particularly for BasisNet, with a threshold of $0.7$, $39.3\%$ of images will skip the second stage thus the average computation cost reduces to 198M MAdds while the overall accuracy remains $80.0\%$ on ImageNet validation set.

\subsection{Convex combination: special cases}
\label{appendix:convex}
\paragraph{Per-model model synthesis.}
When lightweight model predicts a single vector of combination coefficients \emph{for all layers}, \ie $\alpha_1 = \alpha_2 = \dots = \alpha_K \in \mathcal{R}^{N}$, it can be seen as a per-model synthesis. Note that per-model synthesis of BasisNet is still different from HydraNets~\cite{teja2018hydranets}, as the branches in HydraNets span across multiple layers and do not fuse in the middle; instead, in BasisNet the convolution kernels are obtained from linear combination for each layer.

We use BasisNet-MV3 with 8 bases and a lightweight model of MV3-small (1.0x224), and share all layers in basis models except for L11-15. Interestingly both per-model BasisNet and per-layer BasisNet have the same performance, $79.6\%$ top-1 accuracy on ImageNet validation set, implying the combination coefficients across layers may have high correlations for BasisNet-MV3. 
We also experiment with BasisNet-MV2 in a similar setting, but it turns out training per-model BasisNet-MV2 is more challenging because the model easily collapses after roughly 30K steps in our multiple attempts. We suspect that training per-model model synthesis is generally more difficult as it has stronger constraints on the basis models, and it may depend on the base architectures (MobileNetV2 or MobileNetV3). 

\vspace{-1em}
\paragraph{Model selection instead of model synthesis.}
When the predicted combination coefficients are one-hot encoded, the model synthesis can be simplified as \emph{model selection}, as only one base will be selected for a particular layer. We experimented with BasisNet-MV3 with 8 bases, and the lightweight model is MV3-small (1.0x128). Basis models share all layers except for L8-15, and the original BasisNet-MV3 has an accuracy of 79.8\% under this setting. After training for 100K steps we froze the lightweight model and transformed the predicted combination coefficients into one-hot embedding, then continued training the basis models. The resulting BasisNet finally achieved 78.5\% accuracy. This is $+0.7\%$ better than post-processing a well-trained BasisNet ($77.8\%$) implying the potential for training model selection end-to-end.
We leave more careful finetuning for the model selection as future work, but emphasize that model selection has potential to further reduce latency in practice from a model loading I/O perspective. 

\subsection{On-device latency measurements}
\label{sec:latency}

\begin{table}[t]
\centering
\setlength\tabcolsep{1.5pt}
\footnotesize
\begin{tabularx}{\linewidth}{
   >{\hsize=1.6\hsize}X 
   >{\hsize=0.8\hsize\centering\arraybackslash}X
   >{\hsize=0.8\hsize\centering\arraybackslash}X 
   >{\hsize=0.8\hsize\centering\arraybackslash}X
}
 \toprule
 \centering{Models} & Top-1 Acc. & MAdds (M) & Latency (ms) \\
 \midrule
 BasisNet-MV3 8-routes & 79.6\% & 281 & 60.6 \\
 BasisNet-MV3 16-routes & 80.3\% & 290 & 62.9 \\
 -- With early termination & 80.0\% & 198 (avg.) & 43.6 (avg.) \\
 \midrule
 MobileNetV3 (1.25x224) & 79.7\% & 356 & 66.3 \\
 MobileNetV3 (1.5x224) & 80.6\% & 489 & 86.2 \\
 CondConv-MobileNetV3 & 79.9\% & 253 & 53.1 \\
 \bottomrule
\end{tabularx}
\caption{Latency measurements on Google Pixel 3XL.}
\label{tbl:latency}
\vspace{-1em}
\end{table}

To validate the practical applicability, we measured the latency of the proposed BasisNet and other baselines on physical mobile device. We choose Google Pixel 3XL and run floating-point models on the big core of the phone's CPU. In Table \ref{tbl:latency} we show that BasisNet can run efficiently on existing mobile device. Our efficiency conclusion drawn from MAdds also applies to real latency. Specifically, MobileNetV3 with 1.25x and 1.5x multipliers have similar accuracy as BasisNet-MV3 with 8 and 16 routes, while the BasisNet has lower latency. 
We also measured the latency for CondConv. Primarily because of the first stage lightweight model, BasisNet without early termination has higher latency than CondConv (62.9ms vs 53.1ms). However, we emphasize that the first stage lightweight model generates better combination coefficients thus improves the top-1 accuracy (80.3\% vs 79.9\%). Besides, the lightweight model generates initial prediction to enable early termination. When early termination is enabled, the average latency for BasisNet reduced significantly to 43.6ms\footnote{With threshold of 0.7 on ImageNet, 39.3\% of images can skip second stage thus the estimated average latency is reduced to $ 0.393 \times 13.7$ms $+ (1 - 0.393) \times 62.9$ms $= 43.6$ms.}, which is much lower than CondConv (53.1ms) while retaining slightly superior accuracy (80.0\% vs 79.9\%). As we described in Sec.~\ref{sec:basis_models}, deploying early termination for CondConv is much more challenging.

\subsection{Understanding the learned BasisNet models}

\begin{figure*}[t]
  \centering
  \begin{minipage}[b]{.25\textwidth}
  \centering
    \includegraphics[width=\textwidth,align=b]{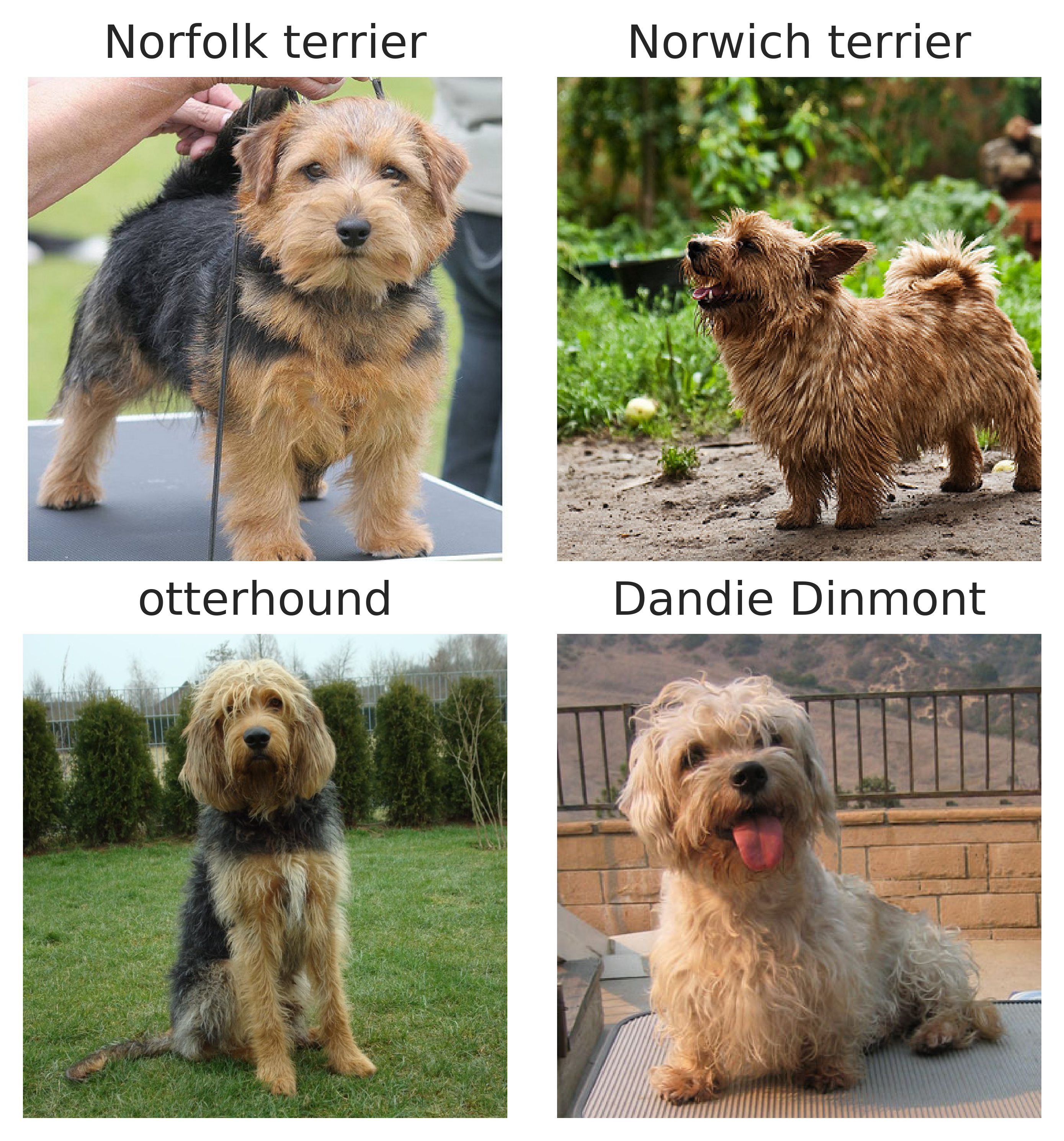}\\
    \footnotesize{(A)}
  \end{minipage}%
  \begin{minipage}[b]{.25\textwidth}
  \centering
    \includegraphics[width=\textwidth,align=b]{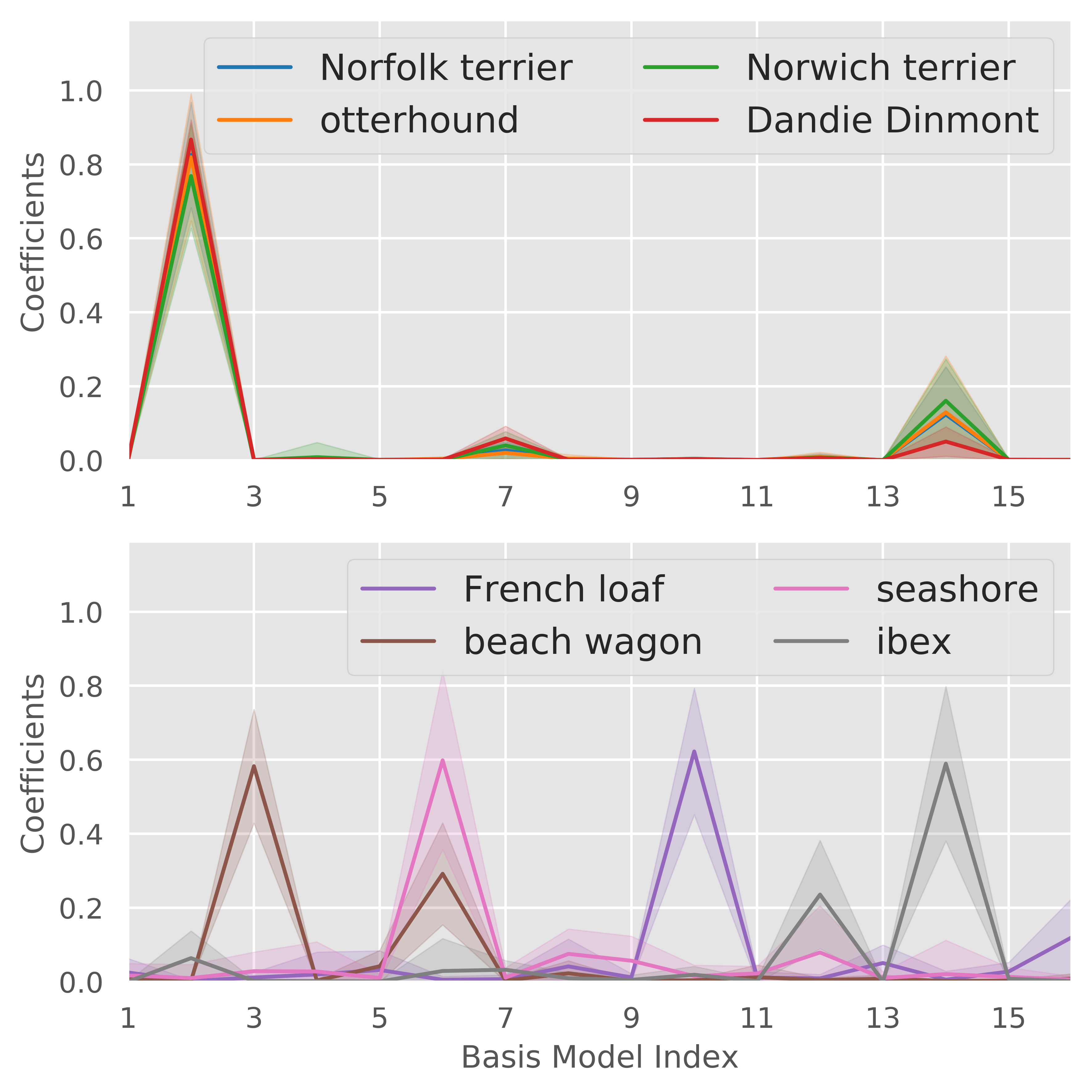}\\
    \footnotesize{(B)}
  \end{minipage}%
  \begin{minipage}[b]{.25\textwidth}
  \centering
    \includegraphics[width=\textwidth,align=b]{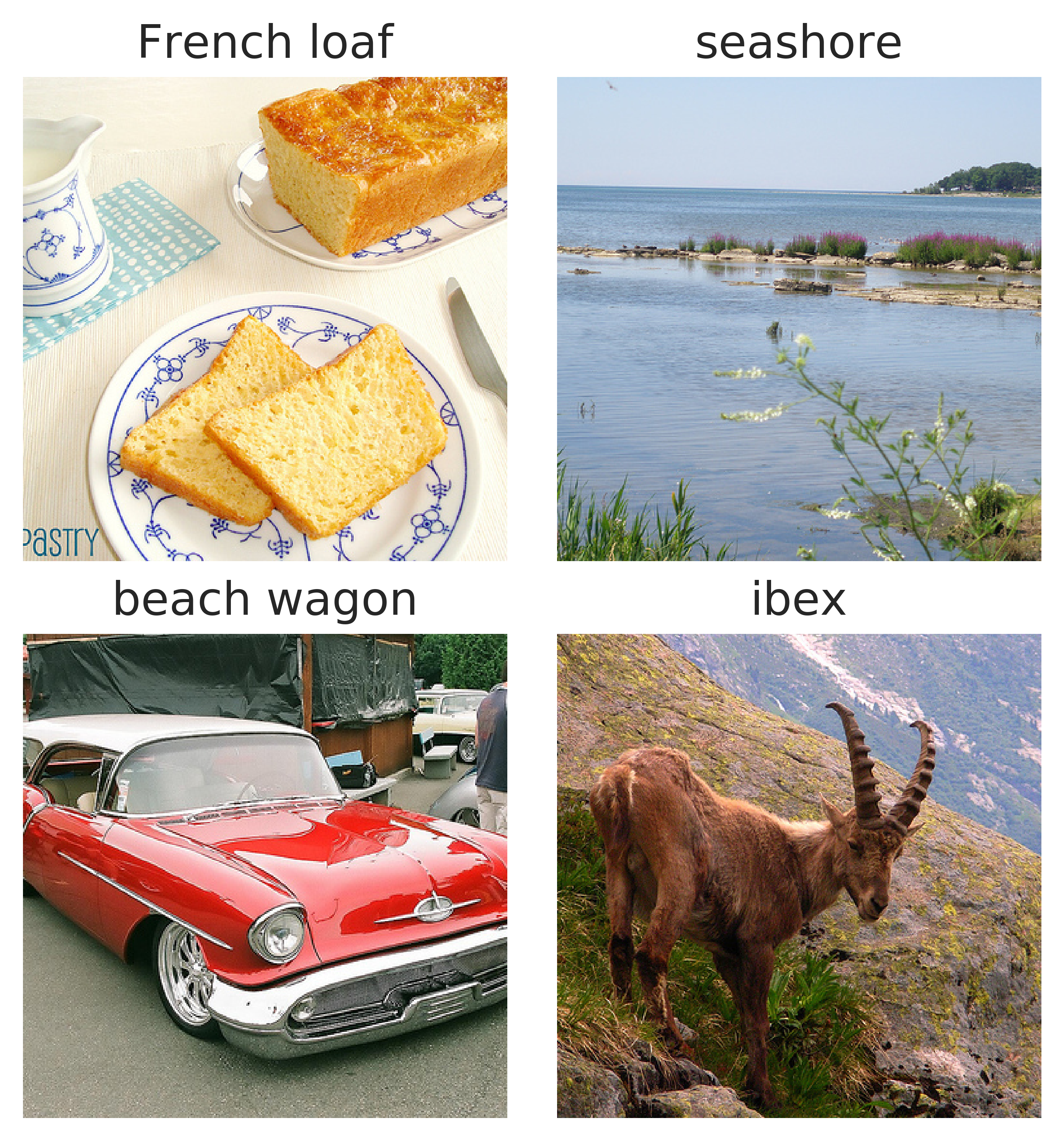}\\
    \footnotesize{(C)}
  \end{minipage}%
  \begin{minipage}[b]{.25\textwidth}
  \centering
    \includegraphics[width=\textwidth,align=b]{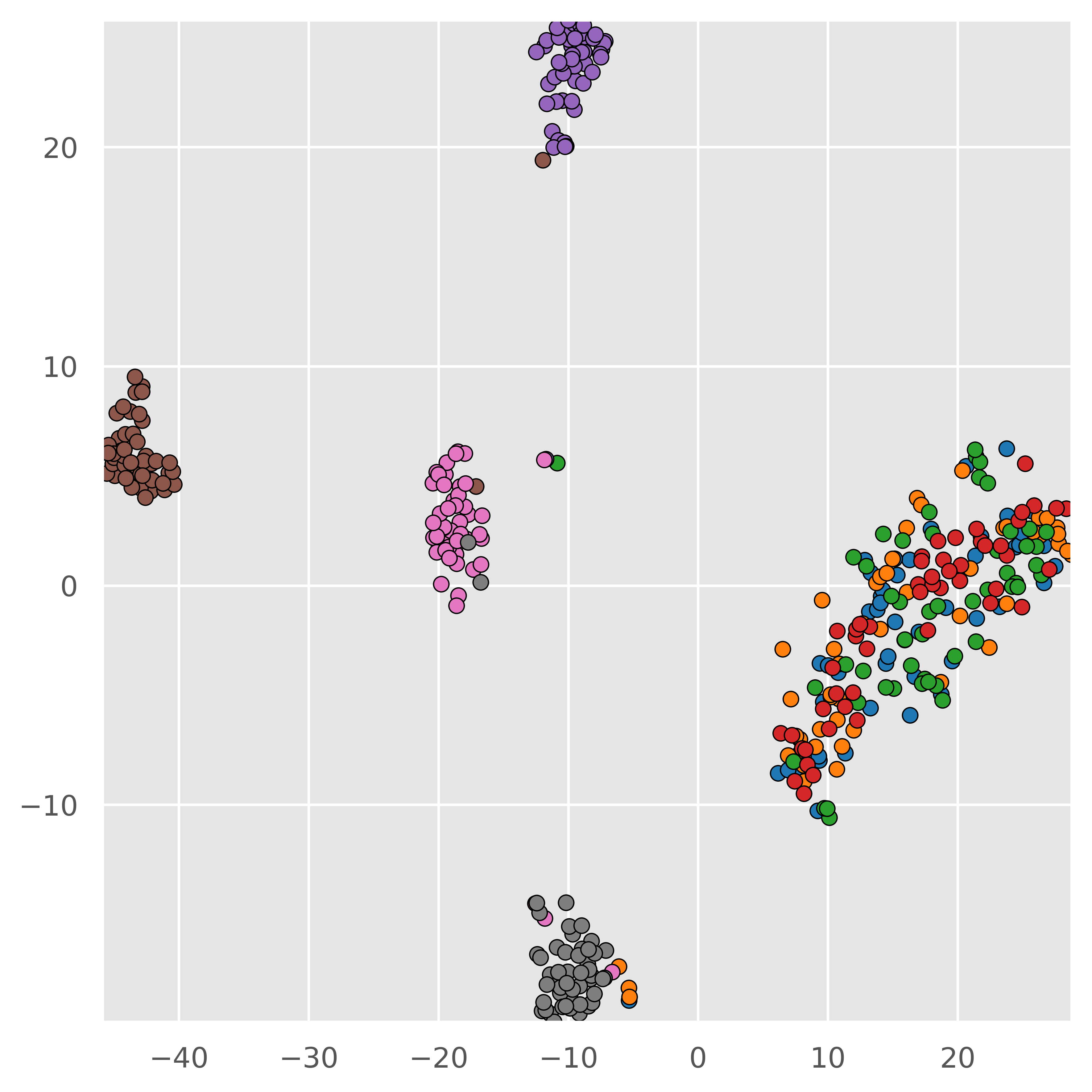}\\
    \footnotesize{(D)}
  \end{minipage}%
\caption{(A,C) Sample images from visually similar or distinct categories. (B) Mean coefficient weights at L15 layer for selected categories. (D) t-SNE visualization of combination coefficients.}
\label{fig:specialization}
\end{figure*}

\paragraph{Visualizing the specialization of basis models.}
We visualized the combination coefficient vectors on ImageNet validation set to better understand the effectiveness of model synthesis. In Fig.~\ref{fig:specialization} we show visually similar and distinct categories, as well as the combination coefficients of L15. 
From (B) top, we can see that the lightweight model chooses the same specialist for distinguishing different dogs, in order to better handle the subtleties between dog breeds. But for visually distinct categories, the synthesized models are very different evidenced by the non-overlapping curves in (B) bottom. In Fig.~\ref{fig:specialization} (D) we show the coefficients for all images using t-SNE~\cite{van2008visualizing}. The dog categories form a single cluster while the others reside in very different clusters. More interestingly, we find that even fine-grained visual patterns can be distinguished as different base models are activated, \eg, fluffy dogs mainly activate 2nd base but short-haired dogs use 14th base. More qualitative results are provided in supplementary materials.

\vspace{-1em}
\paragraph{The importance of optimal basis model synthesis.}
\label{sec:disturbance}

\begin{table}[t]
\small
\centering
\begin{tabularx}{\linewidth}{
   >{\hsize=1\hsize\centering\arraybackslash}X 
   >{\hsize=1\hsize\centering\arraybackslash}X 
   >{\hsize=1\hsize\centering\arraybackslash}X }
 \toprule
 Disturbance & BasisNet-MV2 & BasisNet-MV3 \\
 \midrule
 \textsc{Correct} & 78.2 & 79.8 \\
 \textsc{Top-1} & 73.9 \diff{(-4.3)} & 77.8 \diff{(-2.0)} \\
 \textsc{Mean} & 67.2 \diff{(-11.0)} & 69.5 \diff{(-10.3)} \\
 \textsc{Uniform} & 67.2 \diff{(-11.0)} & 69.7 \diff{(-10.1)} \\
 \textsc{Shuffled} & 56.5 \diff{(-21.7)} & 58.1 \diff{(-21.7)} \\
 \bottomrule
\end{tabularx}
\caption{\label{tbl:disturbance}Apply various disturbance to combination coefficients.}
\vspace{-1em}
\end{table}
To verify the importance of model synthesis, we apply disturbances to the predicted combination coefficients. The specialist should be most effective for the corresponding image, and a disturbed synthesis signal is expected to hurt performance.
We train BasisNet-MV2 (Accuracy $78.2\%$) and BasisNet-MV3 (Accuracy $79.8\%$), and share only the first 7 layers in the basis, then disturb the coefficients $\alpha$ as follows: 
(1) preserving the highest probable basis model only (\textsc{Top-1}), (2) uniformly combining all basis models (\textsc{Uniform}), (3) using mean weights over entire validation set (\textsc{Mean}), or (4) randomly shuffling the coefficients within each layer (\textsc{Shuffled}).
As shown in Table~\ref{tbl:disturbance}, all disturbances lead to inferior performance validating that basis models have varied expertise. \textsc{Shuffled} leads to a totally mismatched specialist thus performance drops over 20 percentage points. 

\vspace{-1em}
\paragraph{Effect of model synthesis at different layers}
\label{appendix:disturbe_layers}

\begin{table}[ht]
\centering
\small
\setlength\tabcolsep{1.5pt}
\begin{tabularx}{\linewidth}{
  >{\hsize=0.8\hsize\centering\arraybackslash}X 
  >{\hsize=1.1\hsize\centering\arraybackslash}X
  >{\hsize=1.1\hsize\centering\arraybackslash}X 
}
\toprule
Disturbed Layer & BasisNet-MV2 & BasisNet-MV3 \\
\midrule
8 & 78.1 \diff{(-0.1)} & 79.6 \diff{(-0.2)} \\
9 & 78.1 \diff{(-0.1)} & 79.6 \diff{(-0.2)} \\
10 & 78.0 \diff{(-0.2)} & 79.6 \diff{(-0.2)} \\
11 & 78.0 \diff{(-0.2)} & 79.4 \diff{(-0.4)} \\
12 & 77.7 \diff{(-0.5)} & 79.0 \diff{(-0.8)} \\
13 & 77.4 \diff{(-0.8)} & 78.3 \diff{(-1.5)} \\
14 & 77.2 \diff{(-1.0)} & 77.9 \diff{(-1.9)} \\
15 & 76.0 \diff{(-2.2)} & 76.4 \diff{(-3.4)} \\
16 & 76.0 \diff{(-2.2)} & 79.1 \diff{(-0.7)} \\
17 & 76.1 \diff{(-2.1)} & N/A \\
18 & 77.2 \diff{(-1.0)} & 76.6 \diff{(-3.2)} \\
\midrule
Reference &  78.2 & 79.8 \\
\bottomrule
\end{tabularx}
\caption{\label{tbl:depth}Top-1 accuracy drops when \textsc{Shuffled} disturbance was applied at different layer. The last row shows the reference model that uses undisturbed predicted coefficients.}
\vspace{-1em}
\end{table}

We also apply disturbances on each individual layer to investigate the sensitivity within the model.
As shown in Table~\ref{tbl:depth}, we find the layers closer to the final classification layer have more impacts, as the accuracy drop is more significant. Interestingly, the regular convolutional layer right after the \textit{residual bottleneck layers}~\cite{sandler2018mobilenetv2,howard2019searching} (e.g. L18 of MobileNetV2 and L16 of MobileNetV3) seems less sensitive towards inputs.

\section{Conclusion}
We present BasisNet, which combines the recent advancements in multiple perspectives such as efficient model design and dynamic inference. With a standalone lightweight model, the unnecessary computation on easy examples can be saved and the information extracted by the lightweight model help synthesizing a specialist network for better prediction. With extensive experiments on ImageNet we show the proposed BasisNet is particularly effective on efficient inference, and BasisNet-MV3 achieves 80.3\% top-1 accuracy with only 290M MAdds even without early termination.

{\small
\bibliographystyle{ieee_fullname}
\bibliography{ref}
}

\clearpage
\appendix

\section*{Supplementary Materials}

\section{Detailed model architecture of BasisNet}
\label{appendix:architecture_details}
Here we describe the details about the proposed BasisNet, including the lightweight model and basis models. 

\subsection{Lightweight model}
\label{appendix:lightweight_model}
For BasisNet-MV2, the lightweight model follows the architecture described in Table~2 of \cite{sandler2018mobilenetv2}, and we use multiplier of $0.5$ and input image resolution of $128$. The lightweight model has a computation overhead of $30.3$M MAdds and a model size of $1.2$M parameters.

For BasisNet-MV3, we use MobileNetV3-small for our lightweight model as described in Table~2 of \cite{howard2019searching}, and we use multiplier of $1.0$ and input resolution of $128$ or $224$ for different experiments. The model size for the lightweight model is $2.5$M parameters regardless of input image resolutions. With $128\times128$ image, the lightweight model has $19.9$M MAdds computation overhead, and with $224\times224$ image the computation overhead is $56.5$M MAdds.

As described in~Sec.~3.1, the lightweight model has two tasks, one for initial classification prediction and the other for combination coefficients prediction. The first task is similar with any regular classification task, and can be formally described as:
\begin{gather}
    \hat{y} = \text{LM}(f(x); W_\text{LM})
\end{gather}
Note that the two tasks share all but the final classification head, thus the extra computation for predicting the combination coefficients is negligible.

\subsection{Detailed architectures for different experiments}
Here we describe the detail about models in different experiments. Unless stated otherwise, we use the following settings as default for BasisNet-MV2 and BasisNet-MV3:
\begin{itemize}[nolistsep,noitemsep]
    \item For MobileNetV2 experiments, the first-stage lightweight model is MobileNetV2 with 0.5x multiplier and input image resolution of 128 (MV2, 0.5x128) and the second stage has 8 basis models of MobileNetV2 1.0x with image resolution of 224 (MV2, 1.0x224). Basis models share parameters in layers from L1 to L10 and final classification layer, and differ in parameters in L11 to L17.
    \item For MobileNetV3, the lightweight model is MobileNetV3-small with 1.0x multiplier and input image resolution of 128 (MV3-small, 1.0x128). The second stage has 16 basis models of MobileNetV3-large with 1.0x multiplier and resolution of 224 (MV3-large, 1.0x224), and they share parameters in first 7 and last 2 layers, and differ in parameters in L8 to L15.
\end{itemize}

\paragraph{Comparison with MobileNets (Sec.~4.2)}
We use BasisNet-MV2 with 8 bases and the lightweight model is MV2 (0.5x128). Each basis model is a MV2 (1.0x224) and they only differ in parameters from L11-17. The basis models dropout rate is $1/8$.

For BasisNet-MV3, we use 16 bases each of a MV3-large (1.0x224), and the lightweight model is MV3-small (1.0x128). All basis models share parameters except for in layers L8-15. The basis model dropout rate is $1/16$.

\paragraph{The effect of regularization for proper training (Sec.~4.3)}
We use the same model architectures for BasisNet-MV2 and BasisNet-MV3 with Sec.~4.2.

\paragraph{Number of bases in basis models (Sec.~4.4)}
We use BasisNet-MV3 with different number of basis models, but each is a MV3-large (1.0x224). The lightweight model is MV3-small (1.0x224) and all basis models share parameters except for layers L11-15. For BasisNet with no more than 8 bases we use basis model dropout rate of $1/8$ and for all others (16 to 128 bases) we use a basis model dropout rate of $1/16$.

\paragraph{Comparison with CondConv (Sec.~4.5)}
For BasisNet-MV3, we use 16 basis models each of a MV3-large (1.0x224), and the lightweight model is MV3-small (1.0x224). All basis models share parameters except for layers L11-15. The basis model dropout rate is $1/16$.

\paragraph{Early stop to reduce average inference cost (Sec.~4.6)}
We use the same BasisNet-MV3 model as in Sec.~4.5.

\section{Implementations and training recipe}
\label{appendix:training_details}
Our project is implemented with TensorFlow~\cite{abadi2016tensorflow}. Following \cite{sandler2018mobilenetv2} and \cite{howard2019searching}, we train all models using synchronous training setup on 8x8 TPU Pod, and we use standard RMSProp optimizer with both decay and momentum set to 0.9. The initial learning rate is set to $0.006$ and linearly warms up within the first 20 epochs. The learning rate decays every $6$ epochs for BasisNet-MV2 ($4.5$ epochs for BasisNet-MV3) by a factor of $0.99$. The total batch size is 16384 (i.e. 128 images per chip). For stabilizing the training, as described in Section~3.3 we keep $\epsilon=1$ for the first 10K training steps then linearly decays to 0 in the next 40K steps. We also used gradients clipping with clip norm of $0.1$ for BasisNet-MV3. In general, all BasisNet and reference baseline models are trained for 400K steps. We set the L2 weight decay to 1e-5, and used the data augmentation policy for ImageNet from AutoAugment~\cite{cubuk2019autoaugment}. We choose the checkpoint from~\cite{xie2019self} as our EfficientNet-b2 teacher model for distillation, and for BasisNet-MV3 both lightweight model and all basis models are trained with teacher supervision. For BasisNet-MV2, we only distill the basis models but use the groundtruths labels without label smoothing for training the lightweight model. For basis models dropout, we use dropout rate of $1/8$ for all BasisNets with no more than 8 bases, and use $1/16$ for the rest which has 16 or more bases. Following \cite{howard2019searching}, we also use exponential moving average with decay 0.9999 and set the dropout keep probability to $0.8$.

\section{Comparison with other efficient networks}
\label{appendix:compare_others}
In Table~1 of our main paper, we show a comparison table with recent efficient neural networks on ImageNet classification benchmark. For baselines we directly use the statistics from the corresponding original papers, even though the training procedures could be very different. Some common tricks in literature include knowledge distillation$\textcolor{capri}{\vardiamondsuit}$, training with extra data$\textcolor{red}{\varheartsuit}$, applying custom data augmentation$\textcolor{black}{\spadesuit}$, or using AutoML-based learned training recipes (hyperparameters)$\textcolor{limegreen}{\clubsuit}$. Different models may choose subsets of these tricks in their training procedure. For example, \cite{xie2019self} use 3.5B weakly labeled images as extra data and use knowledge distillation to iteratively train better student models. CondConv~\cite{yang2019condconv} use AutoAugment~\cite{cubuk2019autoaugment} and mixup~\cite{zhang2017mixup} as custom data augmentation. \cite{wei2020circumventing} reported in a concurrent work that combining AutoAugment and knowledge distillation can have even stronger performance boost, because soft-labels from knowledge distillation helps alleviating label misalignment during aggressive data augmentation. In FBNetV3~\cite{dai2020fbnetv3} the training hyperparameters are treated as components in the search space and are obtained from AutoML-based joint architecture-recipe search. OFA~\cite{cai2019once} use the largest model as teacher to perform knowledge distillation to improve the smaller models. Notably, in our main paper, unless stated otherwise, we \emph{always reported the statistics from our re-implementations}, thus the comparison in our ablation studies are fair, but some results might be inconsistent with this table. It is also worth mentioning that even though we did not explicitly use extra data for training BasisNet, the teacher model checkpoint that we used for knowledge distillation is from noisy student training~\cite{xie2019self}, thus our model may indirectly benefit from the extra data (thus noted by $\textcolor{red}{\heartsuit}$). However, we also experimented with 1.4x MobileNet-V2 as teacher model (which is not exposed to extra data) to train BasisNet-MV2, and verified that the main conclusion still holds.

\section{More quantitative experiments}

\subsection{Detailed comparison with MobileNets (Sec.~4.2)}

\begin{table*}[ht]
\begin{center}
\begin{tabular}{@{}lccccccc@{}}
\toprule
\textbf{Model}        & \textbf{Preprocess} & \textbf{Distillation} & \textbf{\# Bases (BMD)} & \textbf{128}     & \textbf{160}     & \textbf{192}     & \textbf{224}     \\
\midrule
MobileNetV2  & regular    & None         & N/A            & 66.6    & 69.5    & 71.5    & 72.9    \\
MobileNetV2  & AA         & None         & N/A            & 67.8    & 70.7    & 72.7    & 73.7    \\
MobileNetV2  & AA         & MV2 1.4x     & N/A            & 68.8    & 71.4    & 72.4    & 73.1    \\
MobileNetV2  & AA         & EfN-b2       & N/A            & 69.8    & 72.6    & 73.8    & 74.9    \\
BasisNet-MV2 & regular    & None         & 8 (0)          & 68.6    & 71.4    & 73.3    & 74.7    \\
BasisNet-MV2 & AA         & None         & 8 (0)          & 70.4    & 72.8    & 74.6    & 75.6    \\
BasisNet-MV2 & regular    & EfN-b2       & 8 (0)          & 71.8    & 74.8    & 76.2    & 77.2    \\
BasisNet-MV2 & regular    & None         & 8 (1/8)        & 69.1    & 71.9    & 73.7    & 75.0    \\
BasisNet-MV2 & AA         & None         & 8 (1/8)        & 70.9    & 73.2    & 75.1    & 75.9    \\
BasisNet-MV2 & AA         & MV2 1.4x     & 8 (1/8)        & 72.3    & 73.8    & 74.7    & 75.4    \\
BasisNet-MV2 & AA         & EfN-b2       & 8 (1/8)        & 73.5    & 75.9    & 77.0    & 78.1    \\ \bottomrule
\end{tabular}
\caption{\label{table:compare}Detailed comparison of BasisNet-MV2 with MobileNetV2.}
\end{center}
\end{table*}

In Table~\ref{table:compare}, we show original data of Fig.~3 of the main paper, so readers can get the exact accuracy numbers more easily.
Specifically, we show the model performance with different regularizations at 4 different image resolutions \texttt{\footnotesize\{128, 160, 192, 224\}} in the last four columns. We compare the data augmentation (Preprocess, \textit{regular} represents the Inception preprocess as in \cite{sandler2018mobilenetv2,howard2019searching}, and \textit{AA} represents AutoAugment from \cite{cubuk2019autoaugment}), distillation with different teachers (MV2 1.4x represents MobileNetV2 with 1.4x multiplier, EfN-b2 represents EfficientNet-b2 model from \cite{xie2019self}), and basis model dropout. 

We experimented with different teacher network to distill the BasisNet. Note that the MobileNetV2 1.4x teacher we used is from \cite{sandler2018mobilenetv2} and has accuracy of $74.9\%$, and our BasisNet achieves even higher accuracy of $75.4\%$ than the teacher.
We also experimented different variations of EfficientNet (b0, b2, b4, b7) and find that models trained with EfficientNet-b2 has the best performance, and using even better teacher network does not bring performance gain to the BasisNet. We suspect this is related to the gap between teacher and student network as reported in \cite{mirzadeh2020improved}.

\subsection{Detailed experiments for number of bases in basis models (Sec.~4.4)}
\label{appendix:num_bases}
\begin{table}[ht]
\small
\centering
\begin{tabular}{lccc}
\toprule
Model & \#MAdds(M) & \#Params(M) & Acc.(\%) \\ 
\midrule
MV3 (1.0x224) & 217 & 5.45 & 77.7 \\
MV3 (1.25x224) & 356 & 8.22 & 79.7 \\
MV3 (1.5x224) & 489 & 11.3 & 80.6 \\
MV3 (2.0x128) & 276 & 19.1 & 79.2 \\
MV3 (2.5x128) & 435 & 29.0 & 80.4 \\
\midrule
\#Bases=1 & 273 & 8.07 & 77.7 \\
\#Bases=2 & 274 & 9.19 & 78.0 \\
\#Bases=4 & 277 & 11.4 & 78.8 \\
\#Bases=8 & 281 & 15.9 & 79.6 \\
\textbf{\#Bases=16} & \textbf{290} & \textbf{24.9} & \textbf{80.3} \\
\#Bases=32 & 308 & 42.8 & 80.5 \\
\#Bases=64 & 344 & 78.6 & 80.7 \\
\#Bases=128 & 416 & 150.3 & 80.9 \\
\bottomrule
\end{tabular}
\caption{\label{table:bases}Comparison of BasisNets with different number of bases.}
\end{table}

In Table~\ref{table:bases}, we present the original data for Fig.~4 of the main paper, so readers can get the exact accuracy numbers more easily. Notably, we find that BasisNet-MV3 with 16 bases is a good balance between model accuracy and computation budget, achieving $80.3\%$ top-1 accuracy with $290M$ Madds. This table also shows that BasisNet technique optimizes MAdds at the expense of model size.

\subsection{Model synthesis with varying sized lightweight model}

\begin{figure}[ht]
  \centering
  \includegraphics[width=\linewidth]{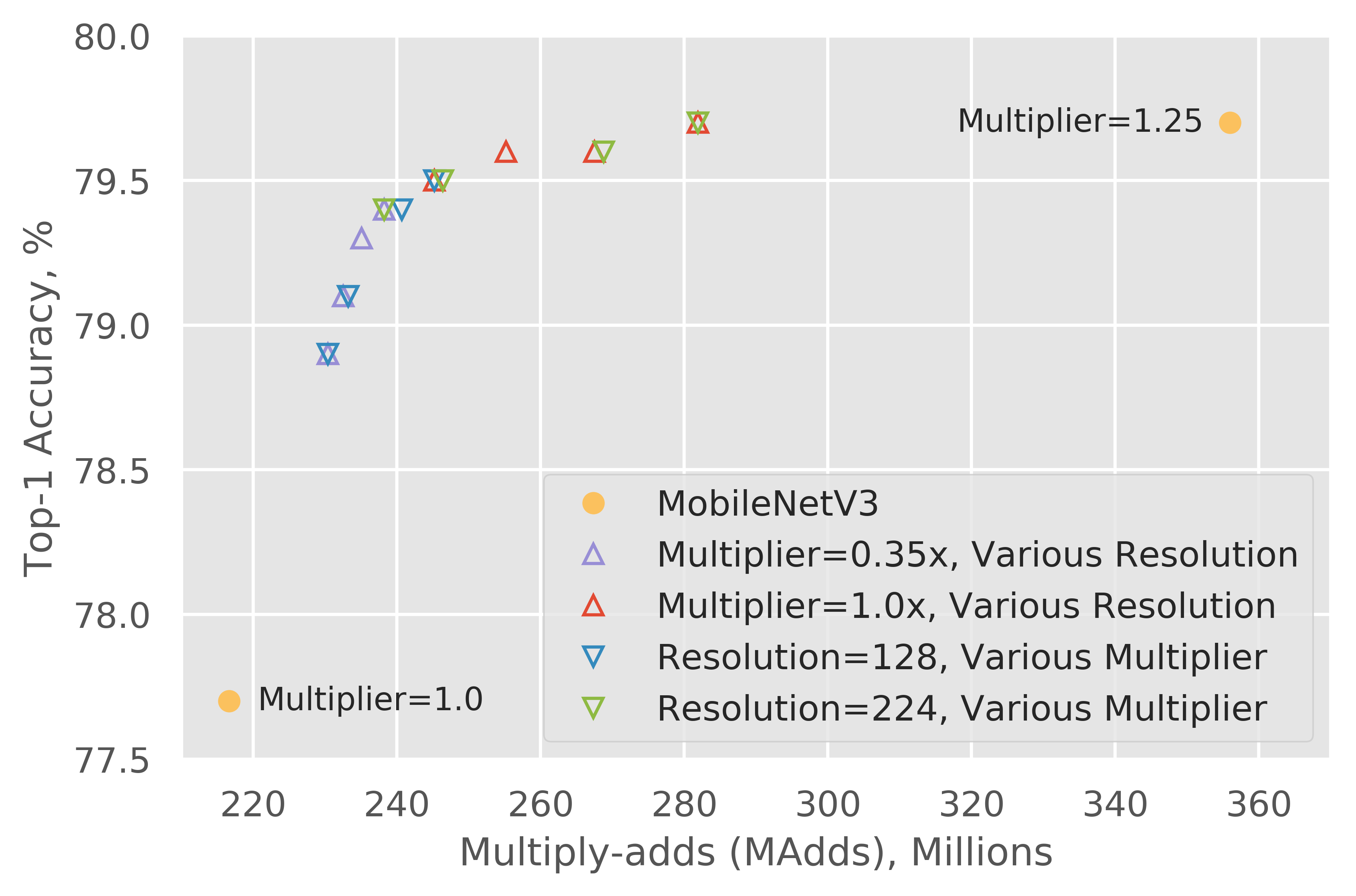}
  \caption{\label{fig:router_size}BasisNet-MV3 with lightweight model of different sizes.}
\end{figure}

We studied the performance of BasisNet with lightweight model of different size. Here the size is measured by the Multiply-adds (MAdds) as we pay more attention to the inference cost. 
We experimented with a BasisNet-MV3 of MV3-large (1.0x224) with 8 bases. The lightweight model is MV3-small, and we experimented with two hyperparameters, i.e. the input image resolution to lightweight model (\texttt{\footnotesize\{128, 160, 192, 224\}}) and the multiplier (\texttt{\footnotesize\{0.35, 0.5, 0.75, 1.0\}}). As shown in Figure~\ref{fig:router_size}, even an extremely efficient lightweight model (MV3-small (0.35x128), computation overhead of $13.8$M Madds) can lead to a performance boost from 77.7\% to 78.9\% (+1.2\%). This experiment shows that resolution and multiplier can have an equivalent effect as reported in \cite{sandler2019non} and a lightweight model with a smaller computation overhead can bring most of the performance gain. Thus it might be more beneficial to scale the model multiplier and resolution coordinately~\cite{tan2019efficientnet}.

\subsection{Model synthesis with early termination}

\begin{figure}[t]
\centering
  \includegraphics[width=\linewidth]{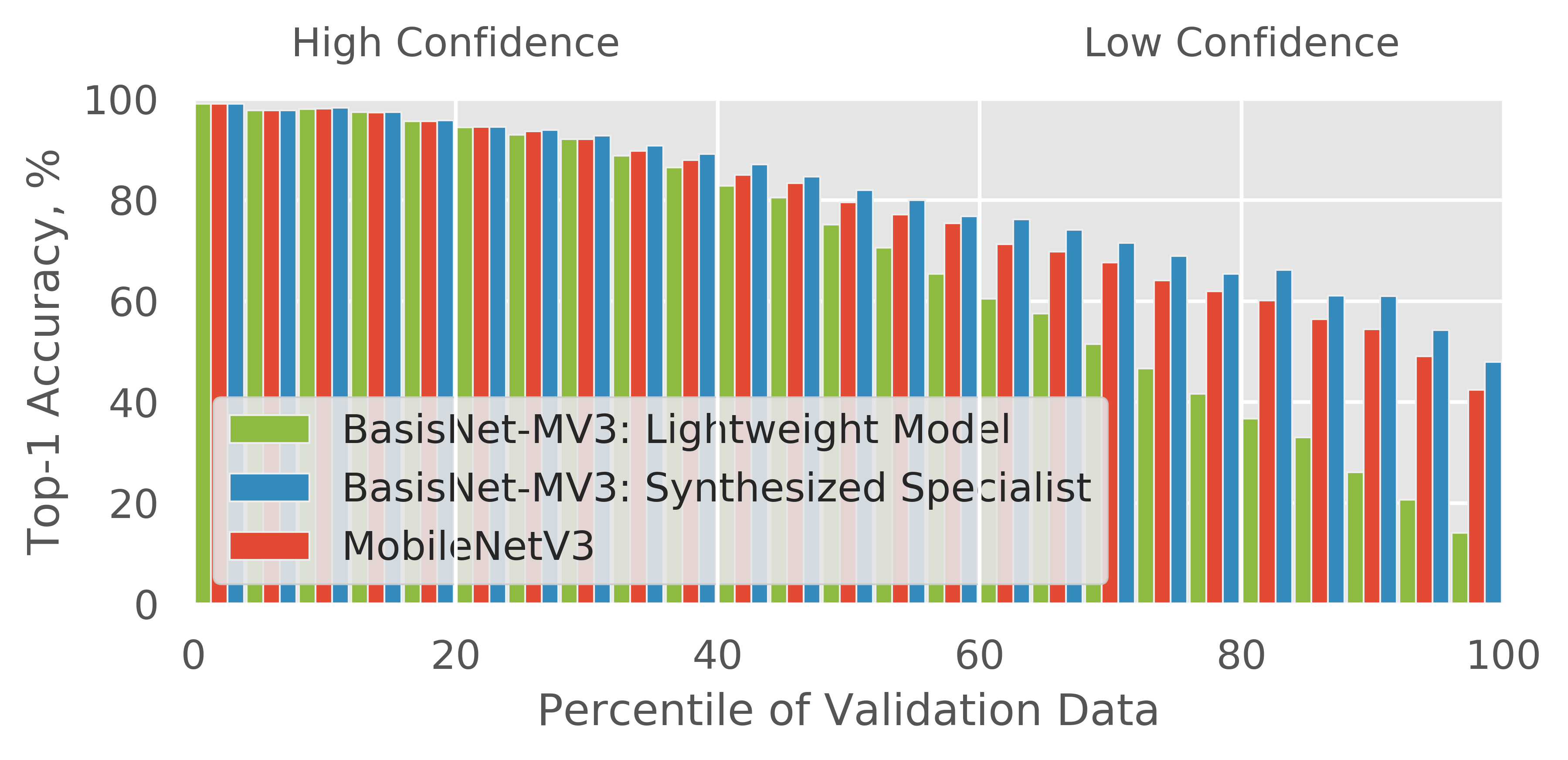}
\caption{\label{fig:sup_early_exiting} Prediction accuracy is comparable for more confident predictions (e.g. top 40\%), and the synthesized specialist consistently outperforms regular MobileNet in all buckets.}
\end{figure}

To better understand the capability of the lightweight model and the synthesized specialist, we split the 50K validation images into multiple buckets according to the sorted highest probability, and show the accuracy of different models for each bucket in Figure~\ref{fig:sup_early_exiting}. Specifically, we show the accuracy within each bucket by the lightweight model, synthesized specialist model and a reference MobileNetV3 baseline. We observe that for at least one third of images where lightweight model has high prediction confidence, the accuracy gaps between these three models are negligible ($<1\%$). The BasisNet has clear advantage over MobileNet in all buckets, especially for more difficult (low confidence) cases.

\section{More qualitative visualizations}
\label{appendix:qualitative}
\subsection{Categories handled by different basis models}
\begin{figure*}[t]
  \centering
  \rotatebox{90}{\scriptsize{\quad Base 2}} \includegraphics[width=0.75\textwidth,align=c]{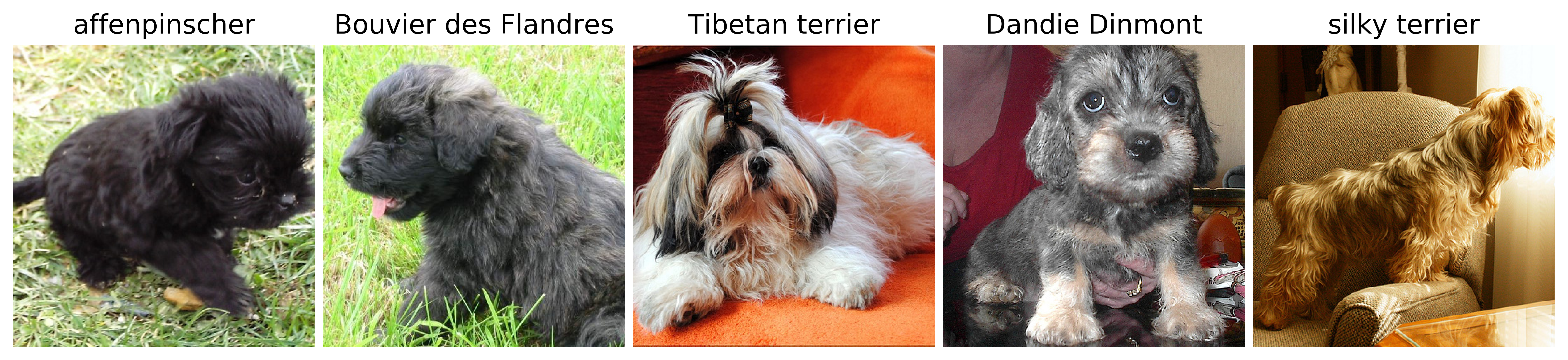}\\
  \rotatebox{90}{\scriptsize{\quad Base 10}} \includegraphics[width=0.75\textwidth,align=c]{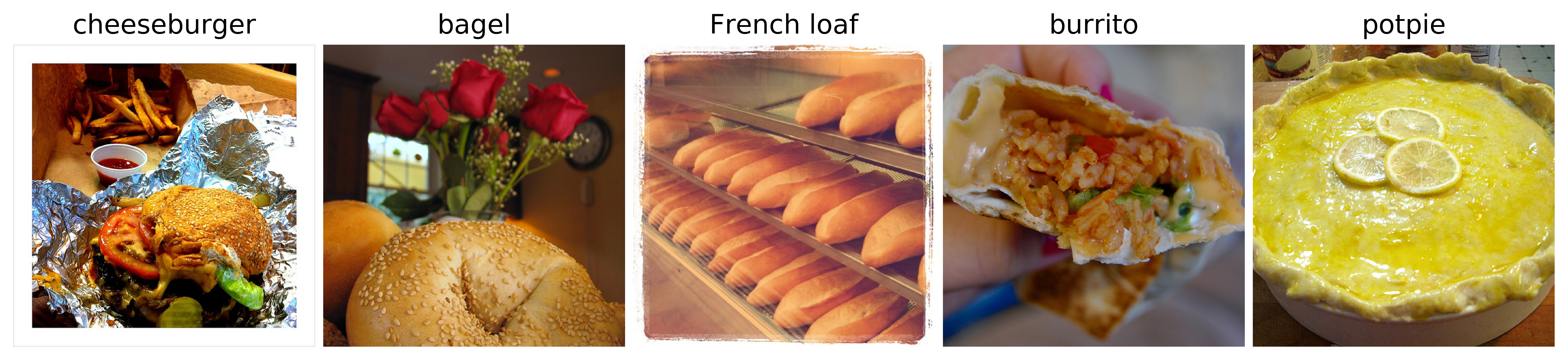}\\
  \rotatebox{90}{\scriptsize{\quad Base 13}} \includegraphics[width=0.75\textwidth,align=c]{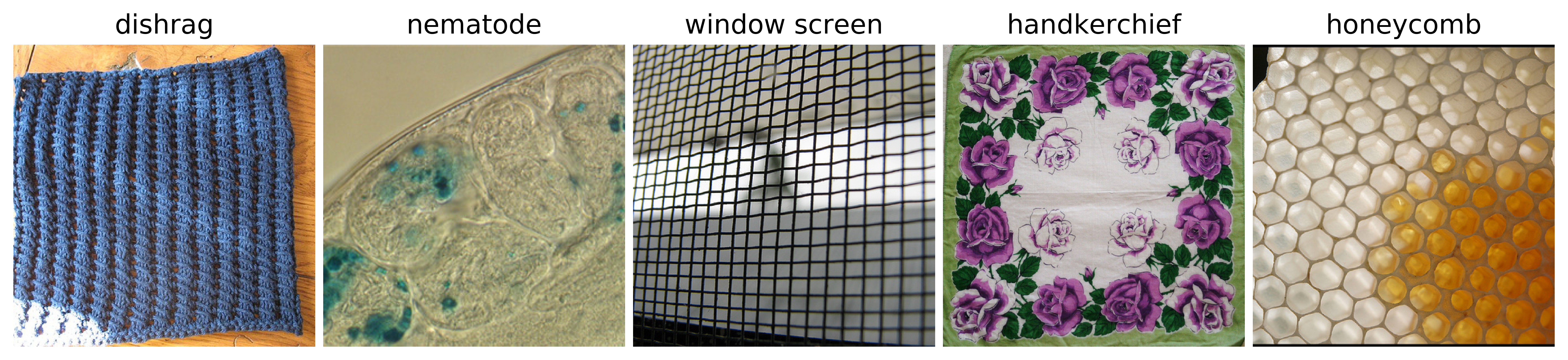}\\
  \rotatebox{90}{\scriptsize{\quad Base 14}} \includegraphics[width=0.75\textwidth,align=c]{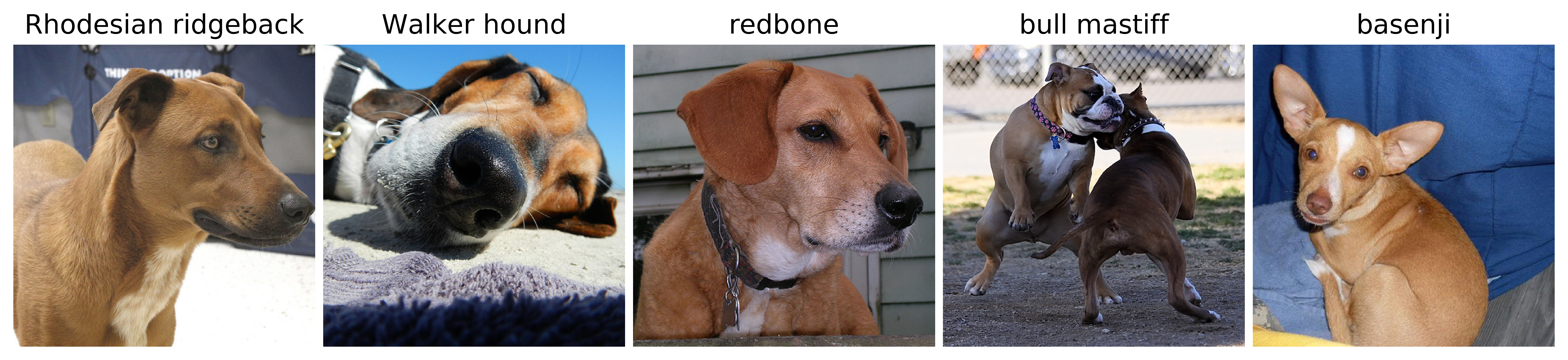}\\
\caption{\label{fig:top_activations}Categories with highest mean coefficients for different basis models.}
\end{figure*}
In Figure~\ref{fig:top_activations} we show several most strongly activated categories for four different basis models on ImageNet validation set. Specifically we trained BasisNet-MV3 with 16 bases, and checked the mean weights at the last non-sharing layer (L15) and show the categories that have the highest mean weights. It is clear that the lightweight model captures the fine-grained visual similarity, for example the base 2 seems to handle the fluffy dogs while the base 14 is more about short-haired dogs. Another example is for base 13 that a clear grid pattern can be found in the images, but semantically these categories are loosely related.

\begin{figure*}[t]
    \centering
    \includegraphics[width=\textwidth]{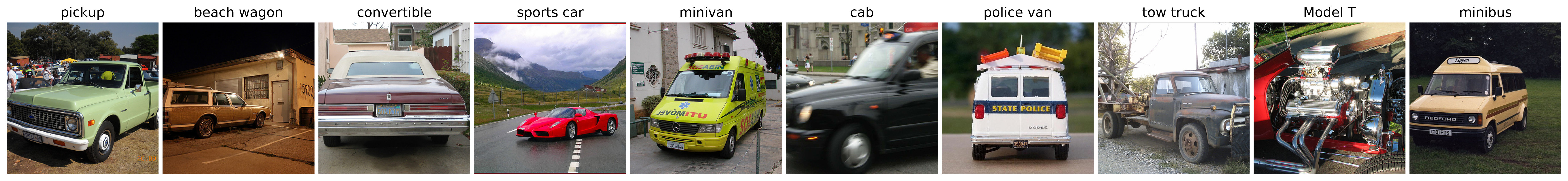}\\
    \includegraphics[width=\textwidth]{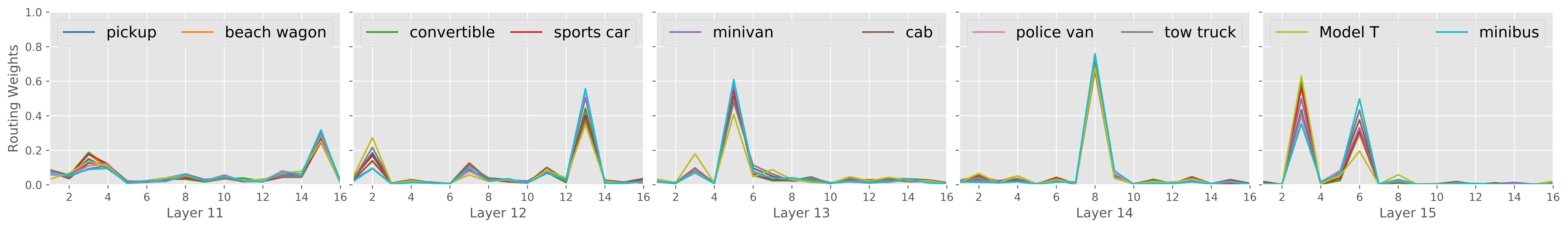}
    \caption{Visualization of predicted combination coefficients for similar categories over all layers.}
    \label{fig:coefficients}
\end{figure*}

\subsection{Combination coefficients for visually similar categories}

In Figure~\ref{fig:coefficients} we show 10 categories regarding different types of cars and the mean predicted combination coefficients for these categories in all layers. Obviously the lightweight model assigns similar coefficients for various cars, implying the effectiveness of the lightweight model. For example, we see that in Layer 14 almost all cars are relying on base 8, and in L15 all cars use a combination of base 3 and base 6. 
Quantitatively BasisNet over these 10 categories have an accuracy of $76.6\%$, but a corresponding regular MobileNetV3 has only $73.2\%$.

\end{document}